\documentclass[10pt,journal,compsoc]{IEEEtran}

\ifCLASSOPTIONcompsoc
  \usepackage[nocompress]{cite}
\else
  \usepackage{cite}
\fi
\usepackage{amsmath,amsfonts}
\usepackage{algorithmic}
\usepackage{algorithm}
\usepackage{array}
\usepackage[caption=false,font=normalsize,labelfont=sf,textfont=sf]{subfig}
\usepackage{textcomp}
\usepackage{stfloats}
\usepackage{url}
\usepackage{verbatim}
\usepackage{graphicx}
\usepackage{cite}
\usepackage{bm}
\usepackage{amssymb}
\usepackage{booktabs}
\usepackage{overpic}
\usepackage[pagebackref,breaklinks,colorlinks]{hyperref}
\usepackage{hyperref}

\usepackage{enumitem}
\usepackage{multirow}
\usepackage{nicefrac}
\newcommand{\paraheading}[1]{\vspace{0.5em}\noindent\textbf{#1}.}

\ifCLASSINFOpdf
 
\else

\fi

\begin{document}
\title{Scalable and High-Quality Neural Implicit Representation for 3D Reconstruction}

\author{Leyuan~Yang,
        Bailin~Deng,~\IEEEmembership{Member,~IEEE},~
        and~Juyong~Zhang$^\dagger$,~\IEEEmembership{Member,~IEEE}
        \IEEEcompsocitemizethanks{
            \IEEEcompsocthanksitem  L. Yang and J. Zhang are with the School of Mathematical Sciences, University of Science and Technology of China.
            \IEEEcompsocthanksitem  B. Deng is with the School of Computer Science and Informatics, Cardiff University.
        }

	\thanks{$^\dagger$Corresponding author. Email: \texttt{juyong@ustc.edu.cn}.}
}

\markboth{~}%
{~}

\IEEEtitleabstractindextext{%
\begin{abstract}
    Various SDF-based neural implicit surface reconstruction methods have been proposed recently, and have demonstrated remarkable modeling capabilities. However, due to the global nature and limited representation ability of a single network, existing methods still suffer from many drawbacks, such as limited accuracy and scale of the reconstruction. In this paper, we propose a versatile, scalable and high-quality neural implicit representation to address these issues. We integrate a divide-and-conquer approach into the neural SDF-based reconstruction. Specifically, we model the object or scene as a fusion of multiple independent local neural SDFs with overlapping regions. The construction of our representation involves three key steps: (1) constructing the distribution and overlap relationship of the local radiance fields based on object structure or data distribution, (2) relative pose registration for adjacent local SDFs, and (3) SDF blending. Thanks to the independent representation of each local region, our approach can not only achieve high-fidelity surface reconstruction, but also enable scalable scene reconstruction. Extensive experimental results demonstrate the effectiveness and practicality of our proposed method.  
\end{abstract}

}

\maketitle

\IEEEdisplaynontitleabstractindextext

\IEEEpeerreviewmaketitle

\IEEEraisesectionheading{\section{Introduction}\label{sec:introduction}}

\IEEEPARstart{R}{econstructing} 3D surfaces from multi-view images is a fundamental problem in computer vision and computer graphics. The recovered dense geometric surfaces provide spatial structure information useful for a variety of downstream applications, such as robotics navigation, animation, physical simulation, virtual and augmented reality, and more.

Traditionally, multi-view stereo (MVS) algorithms~\cite{5226635,609462,10.1007/s00138-011-0346-8} have been one of the preferred methods in the realm of 3D reconstruction. However, these classical algorithms typically demand substantial computational resources and time, and their capability to precisely reconstruct geometric surfaces is constrained. Additionally, a critical limitation of these approaches is their inability to effectively address ambiguous areas within the scene. Such areas encompass repeated texture patterns, extensive regions characterized by uniform colors and low-texture details, semi-transparent elements like fog, or regions with abrupt color transitions. These challenges will further lead to inaccuracies in the reconstruction process, resulting in missing surfaces or significant noises.

Recently, neural radiance fields (NeRF)~\cite{nerf} have been proposed to utilize multi-layer perceptrons (MLP) for representing 3D scenes and achieve impressive results for novel view synthesis. Since the introduction of NeRF, various neural implicit surface reconstruction methods have been developed, and it has become a promising alternative paradigm for reconstruction. These implicit methods utilize a neural network to represent a function encoding the surface, such as the signed distance function (SDF) to the surface~\cite{volsdf, neus}. Thanks to its capability to model complex shapes without being limited by resolution, this new approach has greatly improved the performance of multi-view 3D reconstruction~\cite{neus2, neuralangelo}.

Nevertheless, existing neural implicit surface reconstruction methods still face several challenges. A significant restriction arises from the limited expressive capacity of a single MLP, which may be insufficient for accurate reconstruction of high-frequency details. In addition, due to an MLP's global nature, optimizing the geometry in one part of a scene may unintentionally alter distant parts and lead to undesirable results. 
Furthermore, reconstructing an entire scene with a single network requires global data including global camera poses for the whole scene, which will become increasingly impractical as the data volume increases. For a very large scene, collecting all the data at once is challenging, and pose estimation algorithms~\cite{sfm,mvs,from_coarse_to_fine} require substantial time and computational resources and may even fail to provide accurate global camera poses.

These challenges significantly hinder SDF-based reconstruction. To address these issues, we propose a novel neural implicit representation that applies the divide-and-conquer strategy to the SDF-based realm. 
Unlike previous works that solely rely on global information for block reconstruction of the color radiance field~\cite{block-nerf,Mega-NERF}, we aim to achieve this strategy from the stage of data collection and processing, thinking that this way is essential to realize the goal of scalable large-scale scene reconstruction. Consequently, automatically and conveniently determining the relative positions between different local radiation fields has become an urgent problem to address. Another distinction lies in our focus on solving the discontinuous jump problem arising from splicing in SDFs, rather than solely addressing blending problem in color spaces.

Specifically, unlike previous methods that only use a single radiance field, we design to model an object or scene as a fusion of multiple neural SDFs with overlapping regions. The distribution of loacl neural radiance fields and the overlapping relationship are constructed based on the structure of the object or the layout of the data, and then each local SDF is reconstructed separately. Afterward, we utilize the overlapping to register the camera poses between adjacent radiance fields, which aligns all the local SDFs into a global coordinate system. Finally, the local neural SDFs with common regions are blended using a softmax-based weighting scheme to avoid discontinuity in the overlap area, producing a global SDF that represents the whole scene. Our proposed representation offers several significant advantages compared with existing methods:
\begin{itemize}[leftmargin=*]
    \item First, using separate MLPs to represent different local regions can narrow down the area expressed by each MLP, enabling us to fully utilize its representational power for high-fidelity reconstruction of the details in the whole scene.
    \item Moreover, the use of multiple independent local neural SDFs offers flexibility that benefits the reconstruction and downstream applications: 
    firstly, each local region can recover camera poses in its own coordinate system, avoiding the failure to obtain global poses due to the enormous volume of data; 
    secondly, at the application level, our representation allows extraction, editing, or other usage of individual or a subset of local SDFs without affecting other components.
    \item Finally, our representation makes the reconstruction process scalable: larger scenes can be easily handled by introducing more additional and independent local SDFs to the existing radiation field, while maintaining the reconstruction quality of the local regions. This provides a practical solution that can reconstruct urban-scale scenes.
\end{itemize}

Extensive experiments demonstrate that our proposed representation achieves high-quality and scalable 3D reconstruction, compared to existing methods using a single neural network. By reconstruction with eight local SDFs, we improve the accuracy of Lego's Chamfer distance by up to $45.6\%$, capturing a higher level of intricacy and detail. Additionally, by simply introducing more independent local neural SDFs, we achieve scalable and high-quality reconstruction with high-fidelity texture map covering an area of up to 1200m$\times$800m. Moreover, with the independence of local SDFs, we can implement object-level editing within the scene.

\section{Related Work}
\label{sec:related}

\paraheading{Neural Implicit Surface Reconstruction}
Compared with explicit reconstruction, such as voxels~\cite{937544,609462} and triangular mesh~\cite{10.1117/12.148710,SAHILLIOGLU2010334}, implicit reconstruction employs an implicit function to represent the surface. Consequently, it is continuous, enabling the extraction of the surface at any resolution. IDR~\cite{idr} reconstructs surfaces with neural networks by representing the geometry as the zero level set of SDF. VolSDF~\cite{volsdf} and NeuS~\cite{neus} use a weight function that involves SDF during the rendering process to make colors and geometry closer, resulting in improved quality compared to traditional MVS methods. Furthermore, some works~\cite{MonoSDF,GeoNeus} introduce several priors to make the reconstructed surface more refined. Building on the impressive work of Instant-NGP~\cite{instant-ngp}, NeuS2~\cite{neus2} introduces hash-encoding to further significantly improve the reconstruction speed. To address the limitations of the hash table on the regularization term, Eikonal loss~\cite{eikonal}, Neuralangelo~\cite{neuralangelo} proposes numerical gradient to further refine surface reconstruction. Our representation is universal for neural SDF reconstruction, which means that these methods can serve as models for any local SDF within our method.

\paraheading{Large-scale Scene Reconstruction}
Scene reconstruction is a long-standing problem in computer vision and graphics, and many early optimization-based methods~\cite{5459148,article,2011Modeling,10.1007/s11263-007-0086-4,7780814,10.1145/1141911.1141964,8578578} have been proposed to solve this problem. While these methods have shown impressive reconstruction performance, they often exhibit artifacts or holes in the reconstructed scenes.
Since the introduction of NeRF, several works, such as pretraining networks to learn scene priors~\cite{NEURIPS2021_01931a69,DBLP:journals/corr/abs-2012-02189,DBLP:journals/corr/abs-2107-12512,DBLP:journals/corr/abs-2102-13090,DBLP:journals/corr/abs-2201-12204,zhang2022nerfusion} and exploring grid representations to accelerate training~\cite{instant-ngp,DBLP:journals/corr/abs-2111-11215,9879963,plenoxels}, have enabled neural radiance fields to reconstruct large scenes. Block-NeRF~\cite{block-nerf} extend NeRF to city-level scenes, dividing the scene into multiple blocks for independent reconstruction and then merging them together. Mega-NeRF~\cite{Mega-NERF} propose to use a 3D distance-based method to cluster training pixels into parts that can be trained separately. Grid-guided NeRF~\cite{Grid-guided-NeRF} introduce a two-branch model that combines NeRF-based methods with feature grid-based methods. Switch-NeRF~\cite{switchnerf} design a gating network to dispatch 3D points to different NeRF sub-networks. In this work, we mainly focus on geometric surface reconstruction and develop a scalable 3D reconstruction method.

\paraheading{NeRF Registration}
To enable future NeRF research to be applicable in complex, city-level scenarios, the registration problem of overlapping NeRFs must be addressed. NeRF2NeRF~\cite{nerf2nerf} is the first work to attempt registering multiple NeRFs, calculating the initial transformation based on manually annotated key points and then refining it using Metropolis-Hastings sampling. BARF~\cite{barf} and its variants~\cite{localtoglobal,10003959} contribute to the registration of camera poses by learning the 3D scene representation from NeRFs. Zero NeRF~\cite{zero-nerf} leverages NeRF to register two sets of images with correspondence, although it still require a global registration method for initialization. These methods suffer from various issues, such as relying on global camera poses or needing for human interaction, which is impractical for large-scale scene with extensive datasets. Our registration method achieves automatic registration based on similarity in a small overlapping area, which provides greater convenience and applicability for large-scale reconstruction.

Based on the above relevant excellent works, and in order to solve mentioned limitations in the existing neural implicit SDF-based methods, we propose a scalable and high-quality representation for 3D geometric surface reconstruction.
\section{Method}
\label{sec:method}

\begin{figure*}[th]
  \centering
    \includegraphics[width=1.0\linewidth]{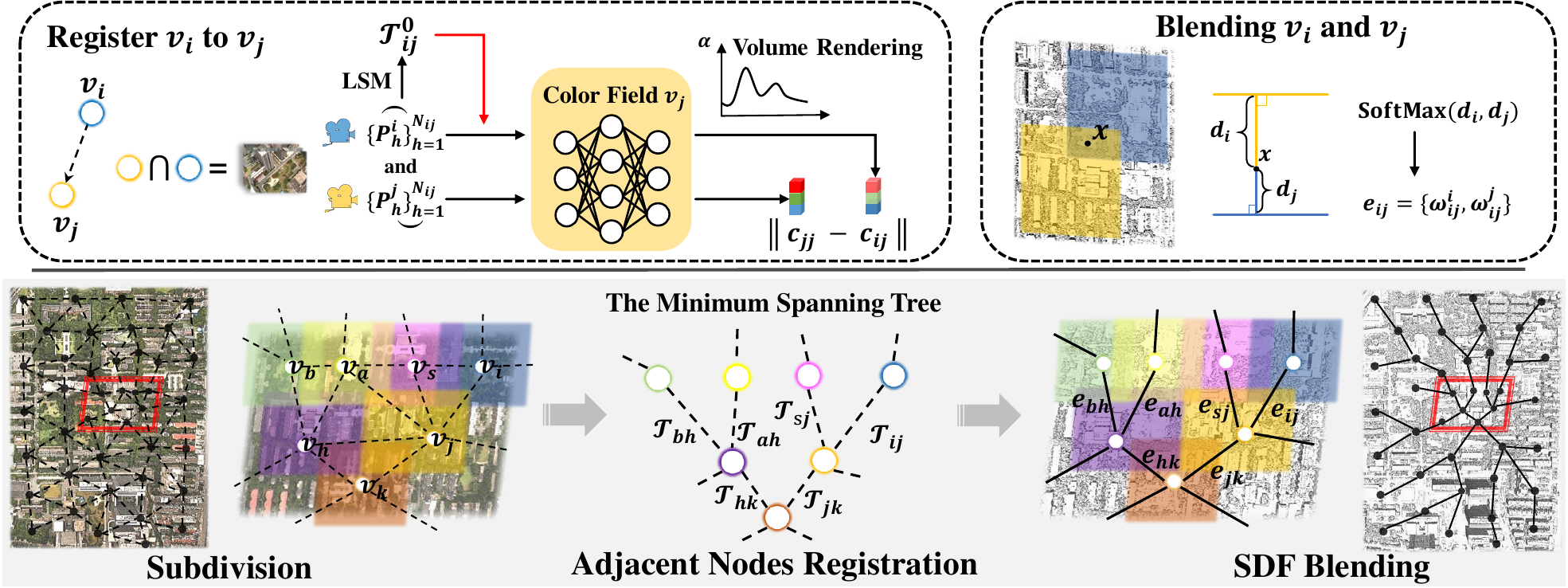}
    \caption{\textbf{Pipeline.} At the bottom, we present the main construction processes of our method: Constructing the distribution and overlap relationship of the local radiance fields based on object structure or data distribution, Adjacent Nodes Registration, and SDF Blending. At the top, we take adjacent nodes $v_i$ and $v_j$ as an example to explain in detail how to use volume rendering for registration and then blending by the softmax-based weighting.}
  \label{fig:pipeline}
\end{figure*}

In the following, we first briefly review the basics of NeRF and neural implicit surfaces in Section~\ref{sec:preliminary}. Afterwards, we present the details of our scalable and high-quality neural implicit representation, including representation (Section~\ref{sec:construct}), registration (Section~\ref{sec:establish}), and blending (Section~\ref{sec:link}). Finally, Section~\ref{sec:application} shows two additional applications based on our method.

\subsection{Preliminary}
\label{sec:preliminary}

\paraheading{Neural volume rendering} Given a set of multi-view posed images, NeRF represents a 3D scene as volume density and color field by volume rendering. Considering a ray $\boldsymbol{r}(t)=\boldsymbol{o}+t\boldsymbol{d}$ emanated from a camera position $\boldsymbol{o}\in\mathbb{R}^3$ in the direction of $\boldsymbol{d}\in\mathbb{R}^3$, the volume rendering scheme essentially involves the integration of radiance from sampled points along the ray. Each 3D sampled point is associated with two spatial factors: volume density $\sigma$ and color radiance $\boldsymbol{c}$, which are predicted using two coordinate-based MLPs. The rendered color of the given pixel is calculated as the integral of the transparency $T(t)=\exp{(-\int_{t_n}^{t}\sigma(\boldsymbol{r}(u))du)}$, density $\sigma(t)$, and radiance $\boldsymbol{c}(t)$ along the ray from the near bound $t_n$ to the far bound $t_f$:
\begin{equation}
  \hat{C} = \int_{t_n}^{t_f} T(t)\sigma(\boldsymbol{r}(t)) \boldsymbol{c}(t) dt.
  \label{eq:volume}
\end{equation}
Practically, this integral is approximated using numerical quadrature.

\paraheading{SDF-based neural implicit surface}
A surface $\mathcal{S}$ can be implicitly represented by the zero level set of its signed distance function $f(\boldsymbol{x}): \mathbb{R}^3 \mapsto \mathbb{R}$, i.e., $\mathcal{S}=\{\boldsymbol{x}\in\mathbb{R}^3|f(\boldsymbol{x})=0\}$. In the context of neural SDFs, NeuS propose converting the volume density in NeRF to SDF representations using a logistic function to allow optimization with neural volume rendering: $$\sigma(\boldsymbol{x})=\phi_s(f(\boldsymbol{x})),$$ where $\phi_s(x)=se^{-sx}/(1+e^{-sx})^2$ is the logistic density distribution which is the derivative of the sigmoid function $\Phi_s(x)=(1+e^{-sx})^{-1}$ with slope $s$. Therefore, an unbiased opaque density function for a ray is:
\begin{equation}
  \sigma(t) = \max 
  \left(
  \frac{-\frac{d\Phi_s}{dt}(f(\boldsymbol{r}(t)))}{\Phi_s(f(\boldsymbol{r}(t)))}
  ,\ 0
  \right).
  \label{eq:sdf}
\end{equation}

\subsection{Representation}
\label{sec:construct}

Unlike existing single radiance field representations, we represent an object or scene as a fusion of multiple radiance fields with overlaps. To this end, we need to first determine the spatial range of each local neural SDF, and subsequently perform local reconstruction for each SDF while ensuring consistency between the SDFs in their overlapping regions.
If we consider each local SDF as a node in a graph, then the overlapping relationship between local SDFs can be interpreted as forming an edge between the corresponding graph nodes. Thus, the representation we propose can be seen as an SDF graph, and we will present it from a graph perspective.

Specifically, we represent a single object or scene as an SDF graph $G=(\mathcal{V},\mathcal{E})$ consisting of a set $\mathcal{V}=\{v_1,...,v_n\}$ of $n$ nodes and a set $\mathcal{E}=\{e_1,...,e_m\}$ of $m$ edges. Each node $v_k$ represents an SDF associated with a local region of the scene, which is reconstructed from a subset of the input images.
For any two nodes $v_i, v_j$ whose image subsets have non-empty intersection, there will be an overlap between their local regions, and we use an edge connecting $v_i$ and $v_j$ to indicate their overlap relation.
In our approach, the local SDF at each node is reconstructed separately, whereas the edges are utilized to register the camera poses from adjacent nodes and align their SDFs, producing a global SDF for the whole scene (see Section~\ref{sec:establish}). To ensure such a global registration can be carried out successfully, the graph $G$ must consist of a single connected component. 

Depending on the structure of the objects and the layout of data, there are different strategies for partitioning the scene to construct an SDF graph.
For instance, if the data is collected based on the components in the scene, such as different parts of a human body shape or different buildings within a campus, then each component can be treated as a node. On the other hand, in some applications, the scene can be divided into grids with overlapping cells. For example, oblique photography datasets can be partitioned based on latitude and longitude. For such cases, each grid cell can be associated with one node. 

After the nodes are determined, we use the image subset associated with each node to derive an SDF for the local region, by utilizing any existing neural implicit surface reconstruction method~\cite{neus, volsdf,MonoSDF,GeoNeus,neus2,neuralangelo}. Thanks to the independence of nodes, we can customize the reconstruction settings for each node to meet specific requirements. For instance, camera poses for each node can be independently estimated. Furthermore, we can select different SDF reconstruction methods for different nodes based on the characteristics of their associated regions. For example, in the case of large scenes, certain locations might demand tailored losses or specialized training strategies to improve reconstruction quality. This level of flexibility is not available in existing methods that rely on a single MLP for the whole scene.

\subsection{Adjacent Nodes Registration}
\label{sec:establish}

Since we only use a subset of the input images to reconstruct the SDF for each node, the resulting SDFs may not share a common world coordinate system across all nodes. Therefore, to derive a global SDF describing the whole scene, we perform registration to align the SDFs with overlapping regions. 

Notably, since an edge in our graph indicates the overlapping relation, a straightforward solution is to perform registration for every edge. However, such a dense registration can be time-consuming and redundant. Instead, we first extract a minimum spanning tree (MST) from the full graph, and then only perform registration on the MST edges. 

Each registration determines a transformation between the coordinate systems for the SDFs on the two nodes connected by an edge. Then, using the  coordinate system of one node as the global coordinate system for the whole scene, we traverse the MST starting from this node, and apply the transformations on the edges to propagate the global coordinate system to all nodes. Afterwards, all local neural SDFs are represented using the global coordinate system. Next, we describe the details of our registration method.

Suppose we perform registration on an edge connecting two nodes 
$v_i$ and $v_j$. We denote the input images and camera poses associated with a node as $\{I^k_1, ..., I^k_{N_k}\}$ and $\{P^k_1, ..., P^k_{N_k}\}$, respectively ($k=i,j$). In addition, let $\{I^{ij}_1, ..., I^{ij}_{N_{ij}}\}$ be the common images between the two nodes, and $\{P^{k}_1, ..., P^{k}_{N_{ij}}\}$ ($k=i,j$) be their corresponding camera poses in the respective coordinate system of each node. For each image $I^{ij}_h$, we use a binary mask $M^{ij}_h$ to indicate the area that appears in the reconstruction results of both nodes. Each camera pose $P$ is represented as $P=[R|T]$, where $R\in SO(3)$ is a rotation matrix and $T\in\mathbb{R}^3$ is a translation vector, is a world-to-camera (w2c) matrix, which denotes the transformation from the world coordinate system into the camera coordinate system:
\begin{equation}
  \boldsymbol{x}_{c} = P \boldsymbol{x}_{w},
  \label{eq:world}
\end{equation}
where $\boldsymbol{x}_w\in\mathbb{R}^3$ is a point in the world coordinate system, and $\boldsymbol{x}_{c}\in\mathbb{R}^3$ is the corresponding point in the camera coordinate system.

\begin{figure}
  \centering
  \includegraphics[width=1.0\linewidth]{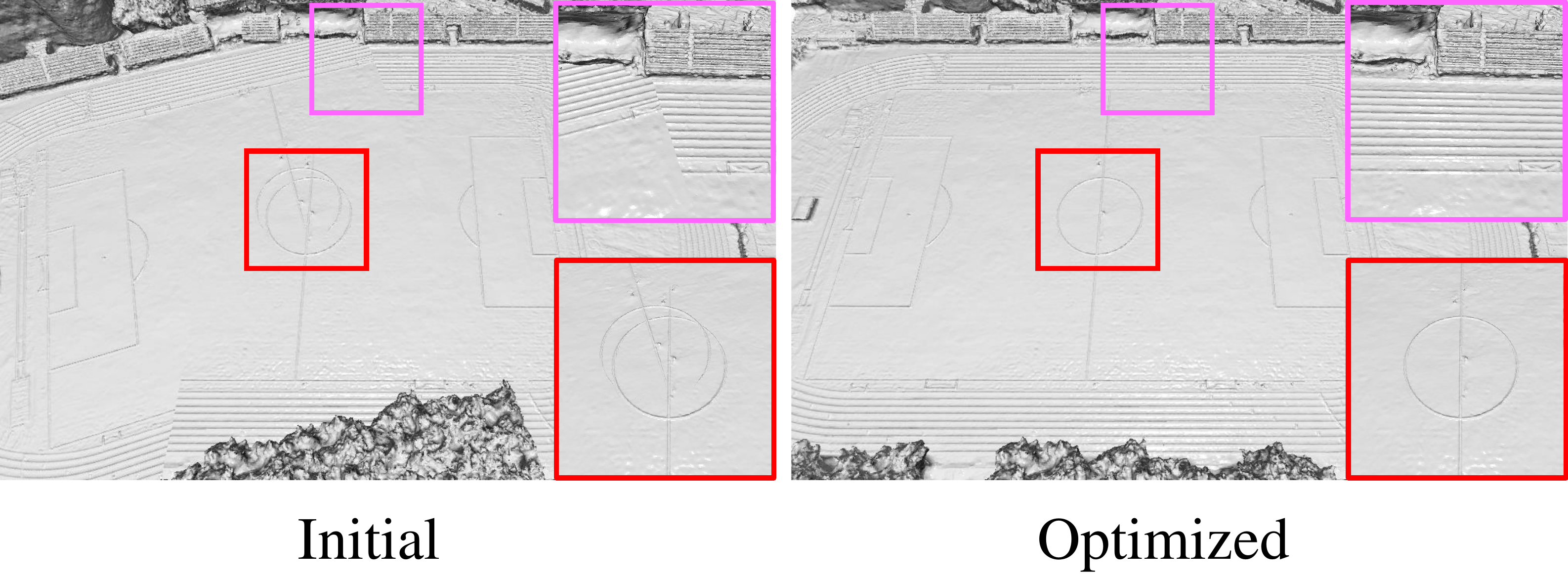}
  \caption{\textbf{Mesh Visualization for Optimization of Registration.} The left image shows misalignment caused by the initial registration, while the right displays the optimized result.}
  \label{fig:ris-ori-opt}
\end{figure}

\paraheading{Initialization for registration}
The purpose of registration is to transform from one world coordinate system to another world coordinate system, corresponding to the training scenes of their respective nodes. We derive the initial registration primarily based on the fact : for any point $\boldsymbol{x}_w$ within the overlapping area of two adjacent nodes, its corresponding point in the camera coordinate system $\boldsymbol{x}_c$ is uniquely determined. 

We denote that point $\boldsymbol{x}_w$ is $\boldsymbol{x}^k_w$ in node $v_k$ ($k=i,j$) in their respective local world coordinate systems. Hence, the corresponding points in the camera coordinate system are obtained after applying the respective w2c matrices for transformation among their common images:
\begin{equation}
    P_h^i \cdot \boldsymbol{x}_{w}^i = P_h^j \cdot \boldsymbol{x}_{w}^j.
    \label{eq:initial-1}
\end{equation}
Additionally, we reevaluate the relationship between corresponding points from a registration perspective. Specifically, we seek the registration transformation $\mathcal{T}_{ij}^0$ for these two adjacent nodes to satisfy:
\begin{equation}
    \boldsymbol{x}_w^i=\mathcal{T}_{ij}^0 \left(\boldsymbol{x}_w^j\right).
    \label{eq:initial-2}
\end{equation}
Therefore, in the local world coordinate system of node $v_i$, through the corresponding camera transformation, there will be:
\begin{equation}
  P_h^i \cdot \boldsymbol{x}_w^i = P_h^i \cdot \mathcal{T}_{ij}^0 \left( \boldsymbol{x}_{w}^j\right).
  \label{eq:initial-3}
\end{equation}
These are two different conversion methods, but the points in the camera coordinate system obtained are uniquely determined. Thus, by observing Eq. (\ref{eq:initial-1}) and Eq. (\ref{eq:initial-3}), we attribute the registration transformation to the conversion of the two local world coordinate systems to the camera coordinate system. Then we should expect the following condition to be satisfied approximately:
\begin{equation}
  P^i_h \cdot \mathcal{T}_{ij}^0= P^j_h,\quad h=1,...,N_{ij}.
  \label{eq:pose}
\end{equation}
Notice that the transformation here becomes right multiplication. More importantly, in this way, the registration transformation can be added to the calculation graph of the optimization network through the step of generating the camera origin and direction, and then can be optimized by volume rendering.

We rearrange these $N_{ij}$ equations to get the following system of equations:
\begin{equation}
    \left(
    \begin{array}{cc}
        R_1^i & T_1^i \\
        R_2^i & T_2^i \\
        \vdots & \vdots \\
        R_{N_{12}}^i & T_{N_{12}}^i
    \end{array}
    \right)
    \cdot
    \mathcal{T}_{ij}^0
    =
    \left(
    \begin{array}{cc}
        R_1^j & T_1^j \\
        R_2^j & T_2^j \\
        \vdots & \vdots \\
        R_{N_{12}}^j & T_{N_{12}}^j
    \end{array}
    \right).
\end{equation}
This leads to an overdetermined system of equations, which can be solved in a least-squares manner. Therefore, we obtain the initial registration transformation expressed in matrix form:
\begin{equation}
  \mathcal{T}_{ij}^0 = 
  \left(
  \begin{array}{cc}
      R_{ij}^0 & T_{ij}^0 \\
      0 & s_{ij}^0
  \end{array}
  \right).
  \label{eq:regis}
\end{equation}
$R_{ij}^0\in SO(3)$ is a rotation matrix, $T_{ij}^0\in\mathbb{R}^3$ is a translation vector, and $s_{ij}^0\in\mathbb{R}$ is a scaling factor. Note that we introduce a scaling factor here, because the SFM system~\cite{sfm} cannot distinguish whether the scales of the world coordinate systems of the two nodes are consistent. Therefore, the transformation between the two world coordinate systems involves not only a rigid transformation, but also a scale expansion. On the other hand, it should be noted that the transformation in Eq. (\ref{eq:initial-2}) will involve the operation of homogeneous coordinates.

\paraheading{Optimization for registration}
It should be noted that the SFM system is not completely accurate. Additionally, the least-squares solver further accumulates and amplifies errors. These factors lead to the inaccuracy of the initial registration matrix, causing misalignment between two nodes after initial registration, as shown in Figure~\ref{fig:ris-ori-opt} and Figure~\ref{fig:rigs-color}. Hence, we further need to optimize the registration matrix to improve the alignment.

It has been demonstrated in~\cite{inerf, parallel} that volume rendering can be used to optimize the intrinsic and extrinsic camera parameters. Our approach to optimizing the registration matrix follows the same principle. However, unlike these works, we use the rendering results within the registered node as supervision rather than ground-truth images.

For the registration of two adjacent nodes, compared to directly optimizing the transformation matrix, we choose to optimize a variation $\Delta \mathcal{T}_{ij}$ with respect to the initial transformation $\mathcal{T}_{ij}^0$. Then the resulting registration is computed as:
\begin{equation}
    \mathcal{T}_{ij} = \mathcal{T}_{ij}^0 + \Delta \mathcal{T}_{ij}.
\end{equation}
Assume the neural radiance fields of both nodes, $\boldsymbol{C}_i$ and $\boldsymbol{C}_j$, have been well trained and fixed. We then use these two poses, the transformed camera pose $P^i_h \cdot \mathcal{T}_{ij}$ and the corresponding pose $P^j_h$ in $v_j$, for volume rendering in $\boldsymbol{C}^j$ and obtain the following colors:
\begin{equation}
    \begin{aligned}
        c^{ij}_h &=\boldsymbol{C}_j (P^i_h \cdot \mathcal{T}_{ij}), \\
        c^{jj}_h &=\boldsymbol{C}_j(P^j_h).
    \end{aligned}
   \quad
  \label{eq:colorsij}
\end{equation}
Theoretically, the colors within the common region indicated by the mask $M_h^{ij}$ should be identical. Therefore, we employ a color loss to optimize the registration matrix:
\begin{equation}
  \Delta \mathcal{T}_{ij}=\arg\min \sum_h \| (c^{ij}_h-c^{jj}_h)\cdot M_h^{ij} \|.
  \label{eq:opti}
\end{equation}

Note that we use the rendering results rather than the ground-truth images as supervision. This is because various factors may hinder accurate learning in certain regions during the training of neural radiance fields. Thus, using the ground truth as supervision may lead to incorrect optimization directions.

In detail, we convert the rotation matrix into an Euler angle representation $E_{ij}^0=(\phi_{ij}^0,\theta_{ij}^0,\psi_{ij}^0)\in\mathbb{R}^3$. Therefore, the variation $\Delta \mathcal{T}_{ij}$ is represented using the change in Euler angles $\Delta E_{ij}$, the change in translation $\Delta T_{ij}$, and the change in scaling factor $\Delta s_{ij}$. Then the resulting registration is derived from the updated Euler angle, translation, and scaling factor computed as:
\begin{equation}
    \begin{aligned}
        E_{ij}&=E_{ij}^0+\Delta E_{ij}, \\
        T_{ij}&=T_{ij}^0+\Delta T_{ij}, \\ 
        s_{ij}&=s_{ij}^0+\Delta s_{ij}.
    \end{aligned}
\end{equation}
In our implementation, we utilize the Adam optimizer~\cite{adam} to solve the optimization problem. The initial learning rate is set to $lr_0 = 5\times 10^{-5}$, and the decay function for the learning rate follows the formula $lr_i= lr_0 \times 0.8^{\frac{i}{100}}$, where $i$ represents the iteration index. We conduct a total of 5K iterations, with $2048$ rays per iteration.

\begin{figure}
    \centering
    \includegraphics[width=1.0\linewidth]{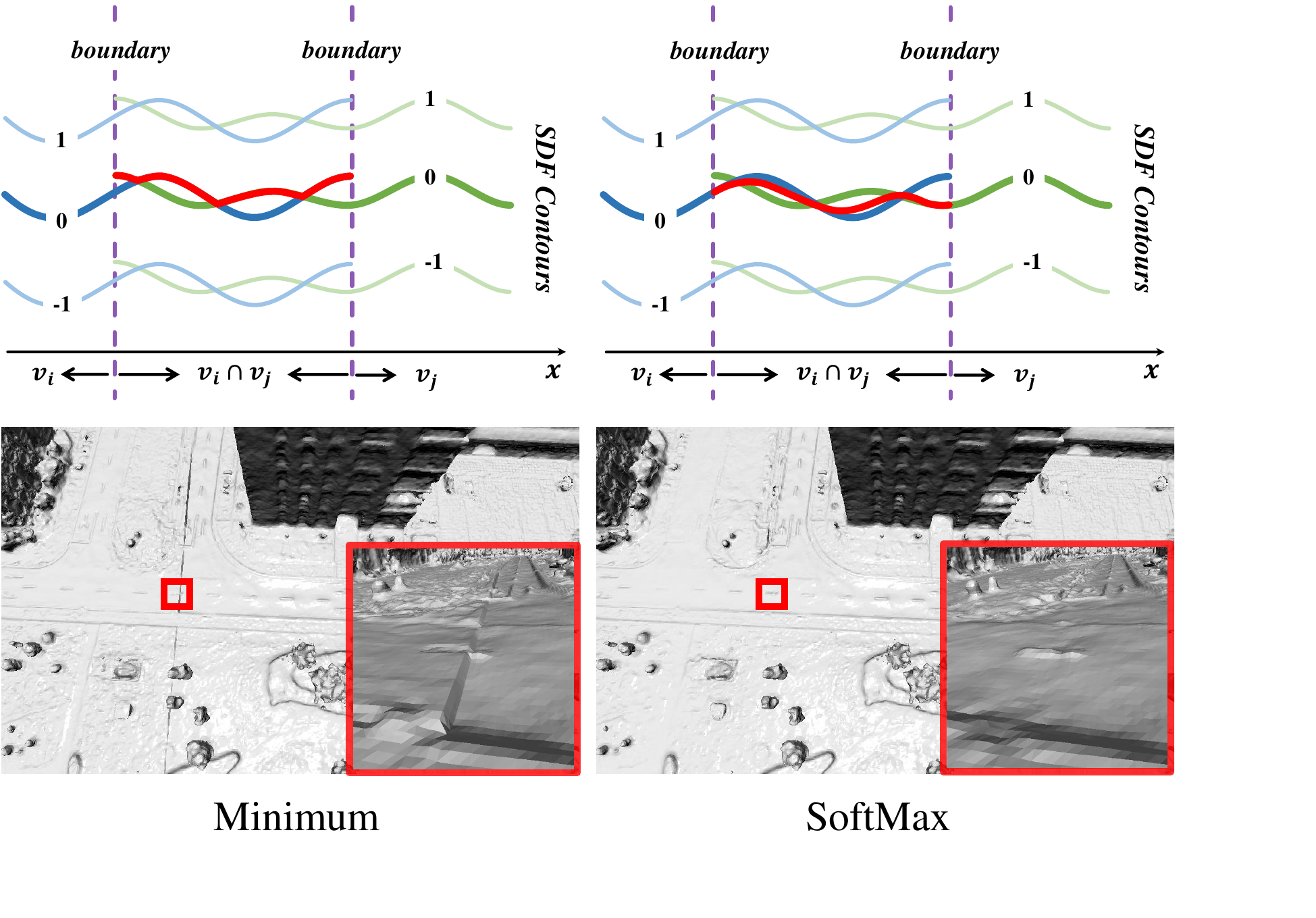}
    \caption{\textbf{Blending.} The top images depict simplified SDF contours within a vertical y-axis section. The left image shows a visible seam caused by directly taking the minimum value, while the right displays the smoothed result achieved through Softmax weighting.}
    \label{fig:blending}
\end{figure}

\subsection{SDF Blending}
\label{sec:link}

After we transform the local SDFs via registration, all SDFs are defined in a common global coordinate system. By construction, these SDFs have overlaps in some areas. However, as reconstruction errors are inevitable, the SDFs may not fully coincide in their overlapping regions. As a result, we need to combine these SDFs in their overlapping regions to derive a consistent global SDF for the whole scene.

Since each SDF represents a part of the scene, one simple solution is to model this as a Boolean union operation that joins the overlapping SDFs. For two objects represented by two SDFs $f_i$ and $f_j$ respectively, it is well known that the object resulting from the union operation is represented by the following SDF~\cite{frisken2006designing,object-nerf,objectsdf++}:
\begin{equation}
    f_{ij}(\boldsymbol{x}) = \min(f_i(\boldsymbol{x}), f_j(\boldsymbol{x})). 
\end{equation}
However, as illustrated in Figure~\ref{fig:blending}, blending the SDFs in this direct manner may result in a noticeable anomalous seam across the boundary of the common region. This occurs due to the abrupt transition between the two local SDFs at the boundary of the overlapping regions, leading to a discontinuity of the blending SDF. To avoid such issues, we instead blend the two SDFs in their overlapping regions:
\begin{equation}
    f_{ij}(\boldsymbol{x}) = \omega_{ij}^i f_i(\boldsymbol{x}) + \omega_{ij}^j f_j(\boldsymbol{x}).
\end{equation}
Here $\omega_{ij}^i$ and $\omega_{ij}^j$ are softmax based weights:
\begin{equation}
  \omega_{ij}^k(\boldsymbol{x})=\frac{ e^{-\beta d_k(\boldsymbol{x})} }{ e^{-\beta d_i(\boldsymbol{x})} + e^{-\beta d_j(\boldsymbol{x})} },\quad k=i,j, 
  \label{eq:weight}
\end{equation}
where $\beta$ is a user-specfied parameter (set to $\beta=10$ in our experiments), and $d_k$ represents the distance from the point $\boldsymbol{x}$ to the boundary of node $v_k$. This scheme increases the relative weight for one SDF as the point  $\boldsymbol{x}$ gets closer to its node boundary, thus smoothing blending the two SDFs and avoiding the undesirable seams (see Figure~\ref{fig:blending}).

After the blending, we obtain a global SDF function for the whole scene, ready for downstream processing such as surface extraction with Marching Cubes~\cite{mc}.

\subsection{Applications}
\label{sec:application}

Our representation offers several benefits compared with existing neural implicit surface reconstruction methods. First, this divide-and-conquer approach utilizes an MLP to reconstruct the SDF in a localized region rather than the whole scene, enhancing its capability to capture high-fidelity details and achieve high-quality surface reconstruction. Second, the graph structure makes the reconstruction process scalable, as we can easily handle larger-scale scenes by adding more nodes while maintaining the computational cost per node and the high-quality of each node. Section~\ref{sec:validation-of-framework-effectiveness},~\ref{sec:precision} and~\ref{sec:large} demonstrate some examples of high-quality and scalable reconstruction using our method.

In addition, the independence of nodes offers flexibility for various downstream applications. In the following, we illustrate the versatility of our method through two additional and practical applications—texture generation and scene editing. Detailed examples of these applications are provided in Section~\ref{sec:tex-and-edit}.

\paraheading{Texturing} 
Our method can not only extract surface geometry but also compute textures for the surface mesh.
To do so, we first extract the corresponding mesh for each node separately and perform parameterization to obtain its texture domain. Then, we initialize the textures separately using the color radiance field within that node. Since the textures are derived from surface rendering and the neural implicit reconstruction relies on volume rendering, there are inherent differences between the two representations, thus the initial texture may contain noises. 
To address this, we employ surface rendering to optimize the texture map of each node. Specially, for node $v_k$, utilizing multi-view images $I_l^k,l=1,...,N_k,$ within this node as supervision, we obtain the rendering images $\hat{I}_l^k$ through rasterization and optimize the texture map using a photometric loss:
\begin{equation}
    \mathcal{L}_{color} = \| \hat{I}_l^k-I_l^k \|.
\end{equation}
Finally, we merge the meshes and textures~\cite{meshlab} from multiple nodes to obtain a large scene's mesh with high-fidelity textures.

\paraheading{Editing}
Another important application of 3D surface reconstruction is the editing of objects within the scene. Through our method, if we consider a single object in the scene as a node, then the independence between nodes provides the flexibility to edit its position in the scene. Specifically, we can adjust the node's registration transformation relative to the global coordinate system, such as adding a certain translation or rotation, and then use our SDF blending to smooth the new boundaries resulting from this modification. In this way, we can manipulate the position and orientation of the objects and move them within the scene.
\section{Experiments}
\label{sec:experiments}

To demonstrate the effectiveness and practicality of our method, we conduct a variety of experiments, including high-quality reconstruction of different categories of objects or scenes, scalable reconstruction of large scenes, and ablation studies. At the same time, we also show its applications in texturing and shape editing.

\subsection{Experimental Configuration and Details}

\paraheading{Datasets} We evaluate our method's capability in high-precision reconstruction using several datasets: 
\begin{itemize}[leftmargin=*]
    \item The NeRF Synthetic Datasets~\cite{nerf} contains rendered RGB images created through Blender, and we use the ``Lego'' object. The Lego dataset contains $100$ images with a resolution of $800$$\times$$800$, as well as corresponding masks and ground-truth camera poses. 
    \item The BlendedMVS datasets~\cite{bmvs} is a large-scale MVS dataset for generalized multi-view stereo networks. Among them, the ``Jade'' object contains 58 multi-view images with a resolution of $768$$\times$$576$, masks, and camera poses. In addition, we chose several large scenes at a higher resolution of images, $2048$$\times$$1536$, for experiments. 
    \item The Actors-HQ Dataset~\cite{humanrf} is a high-fidelity dataset of clothed humans in motion. We use one subject for evaluation, extracting the first frame from each video to create a multi-view image dataset. Thus, one of the human body datasets comprises $160$ images, each with a resolution of $4090$$\times$$2992$ or $2992$$\times$$4090$, accompanied by corresponding camera poses and masks. 
    \item We also created a large-scale, named \emph{Sub-Campus} in the following, which primarily consists of four adjacent buildings in a campus. We used a DJI M300 RTK drone equipped with a Zenmuse P1 camera to capture a total of $2273$ images using oblique photography. These images were then classified based on the different buildings photographed, resulting in four sub-datasets. The corresponding number of images for each sub-dataset is presented in Table~\ref{tab:object-based}. The original resolution of the images is $8192$$\times$$5460$, and  we performed down-sampling to reduce the image resolution to $1024\times682$ for reconstruction.
\end{itemize}
In addition, for the evaluation of scalable large scene reconstruction, we constructed a larger dataset called \emph{Campus} using the same equipment as the Sub-Campus dataset. This dataset covers a larger campus area of about $1200$$\times$$800$~m$^2$ and comprises $5973$ images obtained through oblique photography. Specially, this dataset was collected progressively, beginning from a corner of the scene and gradually expanding to cover the entire area block by block, the specific process as shown in Figure~\ref{fig:scalable-large-scene-process}. Then, based on the division made during collection, we partitioned the entire dataset into $25$ sub-datasets. Similar to the previous Sub-Campus dataset, the captured images were down-sampled from $8192$$\times$$5460$ to $1024$$\times$$682$ resolution.

\begin{figure*}
    \centering
    \includegraphics[width=1.0\linewidth]{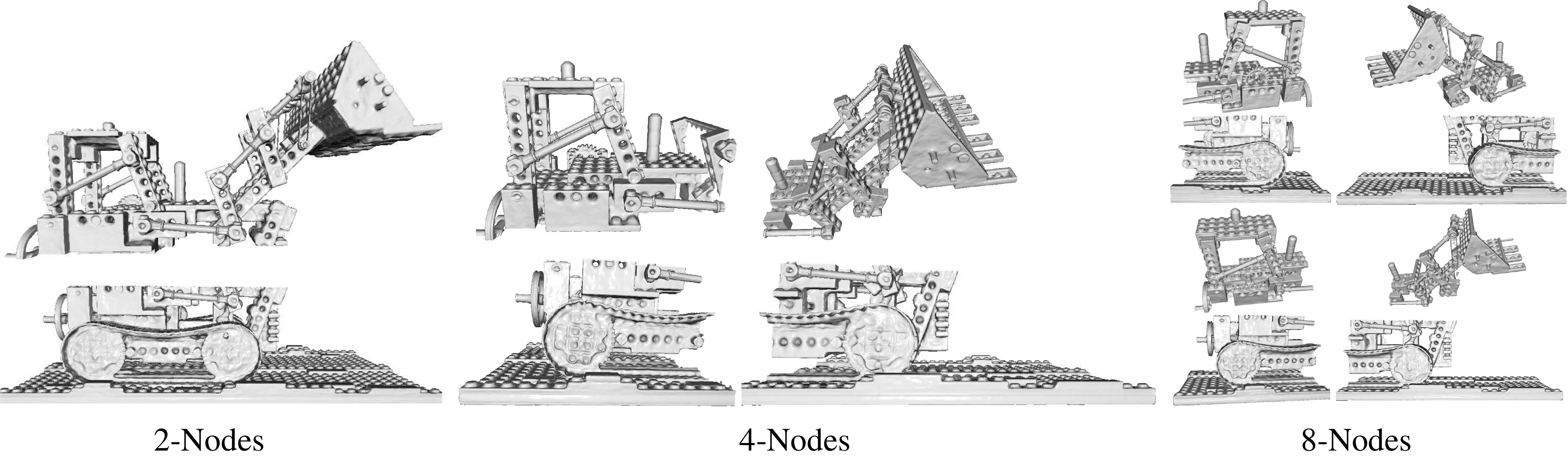}
    \caption{\textbf{Lego Nodes Division.} From left to right are the node divisions of $2$-nodes, $4$-nodes and $8$-nodes.}
    \label{fig:lego-division}
\end{figure*}

\begin{figure*}
  \centering
  \includegraphics[width=1.0\linewidth]{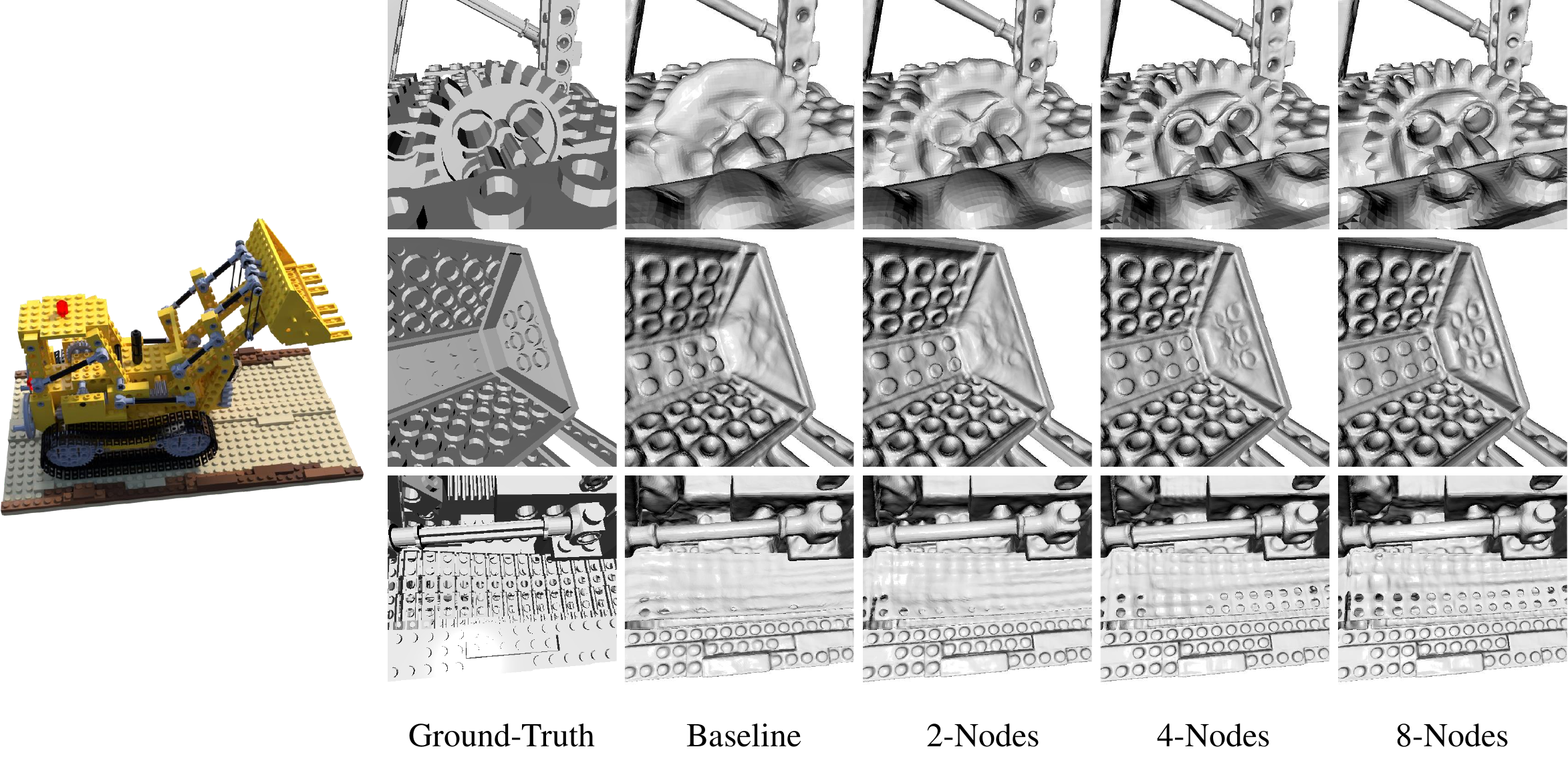}
  \caption{\textbf{Visualization of High-Quality Reconstruction Results for Lego.} The right side is the comparison, demonstrating that our representation progressively reconstructs finer details.}
  \label{fig:lego}
\end{figure*}

\begin{table}
  \centering
  \caption{\textbf{Quantitative Evaluation on the Lego Dataset.} CD represents the Chamfer distance from the ground-truth mesh to the reconstructed mesh. The last column represents the mean of the SDF absolute values corresponding to points uniformly sampled on the ground-truth surface. The optimal result is indicated in bold.}
  \begin{tabular}{@{}lccc@{}}
    \toprule
     & CD$\downarrow$ ($10^{-4}$) & F-score$\uparrow$ (threshold=$5e^{-4}$) & SDF$\downarrow$ ($10^{-3}$) \\
    \midrule
    Baseline & 5.209 & 0.8445 & 8.779 \\
    2-nodes & 4.743 & 0.8621 & 7.348 \\
    4-nodes & 3.877 & 0.8826 & 7.054\\
    8-nodes & \textbf{2.836} & \textbf{0.8972} & \textbf{6.749}\\
    \bottomrule
  \end{tabular}
  \label{tab:lego}
\end{table}

\paraheading{Evaluation metrics} 
We calculate the Chamfer distance~\cite{DBLP:journals/corr/QiSMG16} and the F-score~\cite{Knapitsch2017} if the ground-truth mesh is available, such as the Lego dataset. For the BlendedMVS dataset, we utilize the provided textured mesh reconstructed using the MVS method~\cite{bmvs} as the ground truth. For the Actors-HQ dataset, we utilize the provided mesh of the first frame as the ground truth, which comes from the frame-by-frame mesh reconstructed by applying Humanrf~\cite{humanrf} at 2x resolution. Additionally, to mitigate the impact of errors introduced by Marching Cubes during mesh extraction, we uniformly sample a specific number of points on the ground-truth mesh and then calculate the average absolute SDF values of these points within the reconstructed SDF. For the large scene datasets Sub-Campus and Campus where no ground-truth mesh is available, we qualitatively visualize and magnify the reconstruction details for evaluation.
Furthermore, we report rendering quality metrics\textemdash{}including~Peak Signal to Noise Ratio (PSNR), Structural Similarity (SSIM)~\cite{SSIM}, and Learned Perceptual Image Patch Similarity (LPIPS)~\cite{LPIPS}\textemdash{}both before and after the registration optimization to illustrate the effectiveness of our approach.

\begin{figure}
  \centering
    \includegraphics[width=1.0\linewidth]{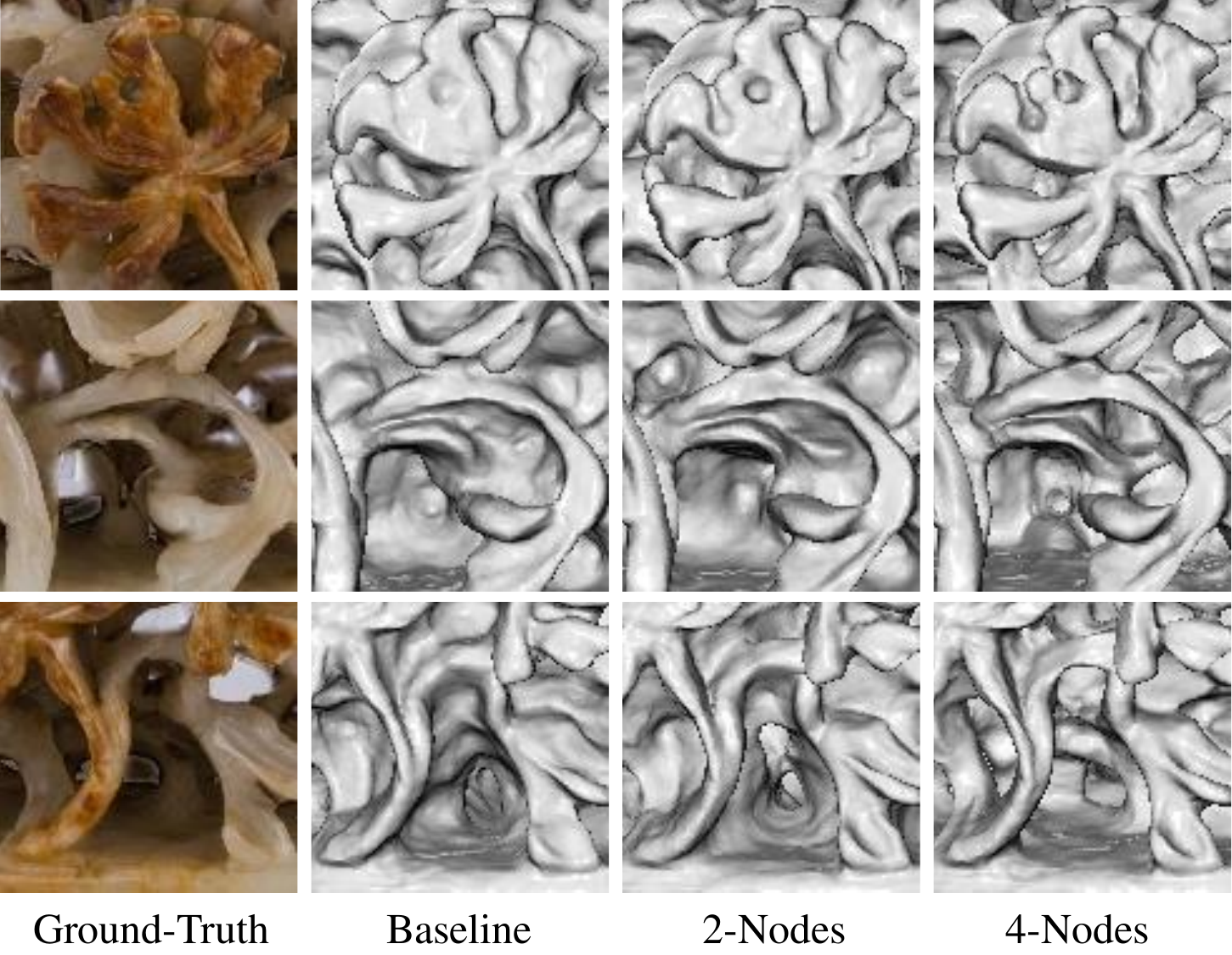}
    \caption{\textbf{Visualization of High-Quality Reconstruction of Jade.} It can be seen that as the number of nodes increases, the carvings and hollowing out of the jade are gradually reconstructed and refined.}
  \label{fig:jade}
\end{figure}

\begin{table}
  \centering
  \caption{\textbf{Quantitative Evaluation on the Jade Dataset.} The optimal result is indicated in bold.}
  \begin{tabular}{@{}lccc@{}}
    \toprule
     & CD$\downarrow$ ($10^{-4}$) & F-score$\uparrow$ (threshold=$5e^{-4}$) & SDF$\downarrow$ ($10^{-3}$) \\
    \midrule
    Baseline & 2.507 & 0.7801 & 8.527 \\
    2-nodes & 2.431 & 0.7847 & 7.548 \\
    4-nodes & \textbf{0.958} & \textbf{0.7873} & \textbf{4.136}\\
    \bottomrule
  \end{tabular}
  \label{tab:jade}
\end{table}

\paraheading{Implementation details} 
We run our experiments on a workstation with an Intel Xeon E5-2690 v3, an NVIDIA GeForce RTX 3090, and 128GB of RAM.  
We use the open-source implementation of NeuS~\cite{neus}\footnote{\url{https://github.com/Totoro97/NeuS}} as well as the open-source project SDFStudio~\cite{SDFStudio}\footnote{\url{https://github.com/autonomousvision/sdfstudio}} for reconstructing SDF at a node and for the whole scene.
Camera poses are estimated using COLMAP~\cite{mvs,sfm}.
Our initial registration is computed using NumPy and takes no more than $1$ second. The registration optimization is performed using PyTorch with $5k$ iterations and $2048$ rays per iteration, and takes about $15$ minutes on average.
More implementation details can be found in the supplementary materials.

\subsection{Validation of Effectiveness}
\label{sec:validation-of-framework-effectiveness}

We aim to first verify the effectiveness of the proposed graph-based SDF representation from two aspects:
\begin{itemize}
    \item Firstly, we demonstrate the inherent limitations in the expressive capacity of a single MLP. In other words, as the complexity of objects or scenes increases, a single MLP may fail to reconstruct refined geometry effectively.
    \item Secondly, we validate the feasibility and effectiveness of node partitioning. Specifically, by narrowing down the spatial range expressed by a single MLP, we can improve the quality and accuracy of reconstruction.
\end{itemize}

\paraheading{Baseline} 
To comprehensively validate these assertions, we employ NeuS~\cite{neus} as our baseline method, representing the most fundamental method within the neural SDF domain. This choice serves two primary purposes: Firstly, NeuS is the foundational representation for subsequent improvements in neural implicit surface reconstruction, and its validation is crucial for affirming the general feasibility of future enhancement efforts. Secondly, NeuS does not rely on any additional prior information or regularization loss during training, thus ensuring that extraneous factors do not influence our experimental verification results. Even though there are some improved SDF models that can achieve better performance than NeuS on the same scene, these single-MLP models still run into the same issues eventually if the scene becomes large and complex enough. Therefore, this issue is universal for single-MLP approaches regardless of the model used.
In the following experiments, we utilize the Lego and Jade datasets as reconstruction targets, representing two different datasets of synthetic and real scenes respectively. Therefore, the baseline is reconstructing the entire object using only a single MLP with NeuS.

\paraheading{Our method} To apply our method, we partition the bounding box along axis directions into overlap cuboids, resulting in a graph with 2, 4, and 8 nodes, respectively, as shown in Figure~\ref{fig:lego-division}. 
For the situation where Lego is divided into upper and lower nodes, the bounding box range on the $z$-axis is $[-0.35, 1.01]$, and we directly divide it into $[-0.35, 0.25]$ and $[0.0, 1.01]$. This uneven division accounts for the fact that most parts of the scene are located at the lower part of the bounding along the $z$-axis. The overlapping areas of the two nodes account for 41.67\%, 24.75\%, and 18.38\% of the lower node range, upper node range, and global range, respectively. 
For the case involving four nodes, we build upon the aforementioned division into two nodes and further partition the bounding box range along the $y$-axis, $[-1.15, 1.15]$, into $[-1.15, 0.4]$ (representing the car rear area) and $[0.0, 1.15]$ (representing the front bucket area). The overlapping area constitutes 25.81\%, 34.78\%, and 17.39\% of the global range along the tail, front, and $y$-axis, respectively. Similar to the division along the $z$-axis, the uneven division here accounts for the distribution of the scene within the bounding box.
For the scenario with eight nodes, we build upon the previous division into four nodes and further partition the bounding box range along the $x$-axis ($[-0.64, 0.64]$) into $[-0.64, 0.10]$ and $[-0.10, 0.64]$. The overlapping area constitutes 27.02\% and 15.63\% of one side and the global range of the $x$-axis, respectively. 
For the Jade dataset, we directly divide its bounding box into overlapping cuboids of the same size along the axes, and the overlapping areas account for 20\% of the global range.
For a fair comparison, we use the same reconstruction model and configuration for the baseline and for each node. Moreover, the reconstructed surfaces are extracted via Marching Cubes using the same settings.

\begin{figure*}
  \centering
    \includegraphics[width=0.49\linewidth]{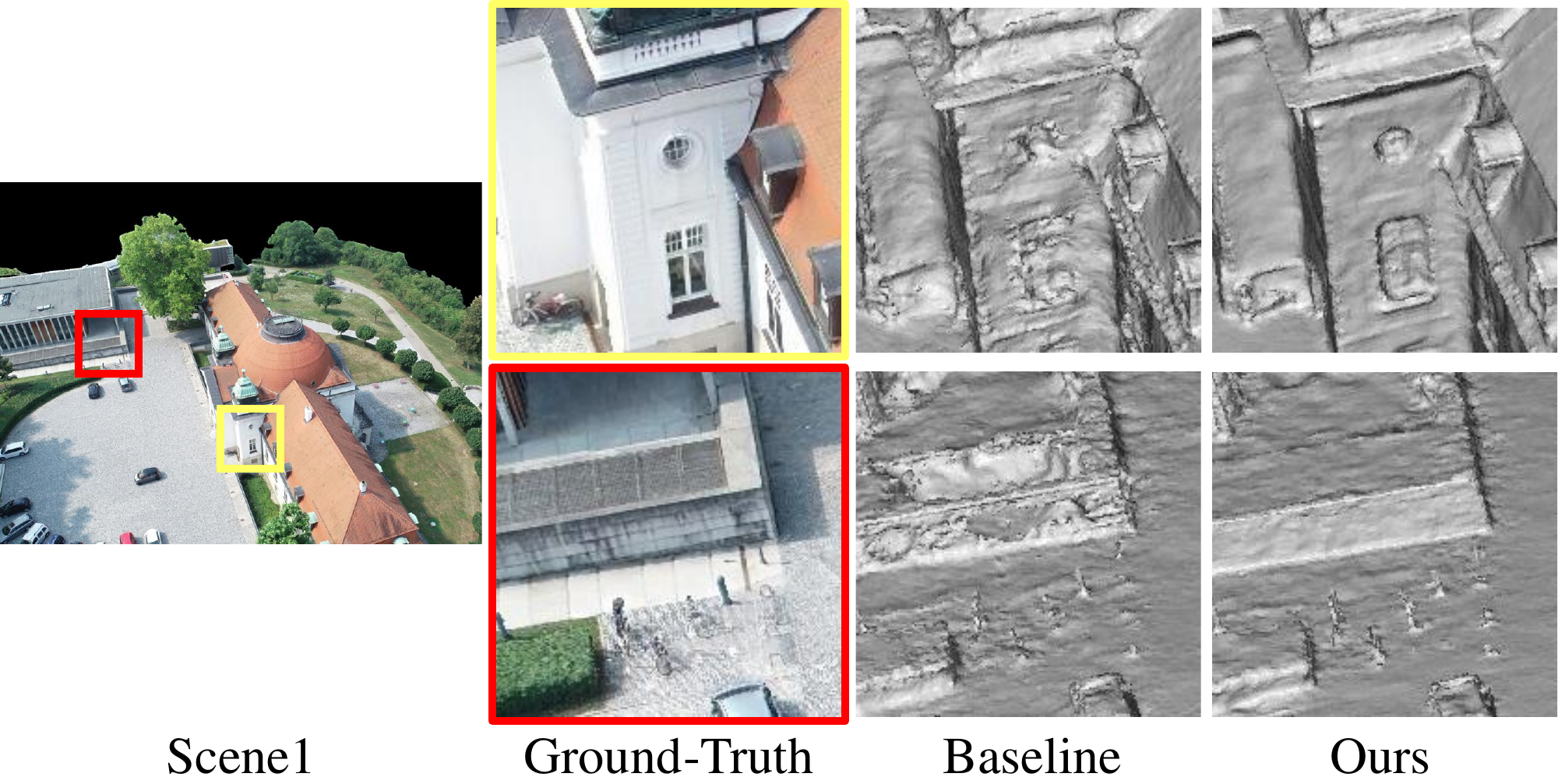}
    \includegraphics[width=0.49\linewidth]{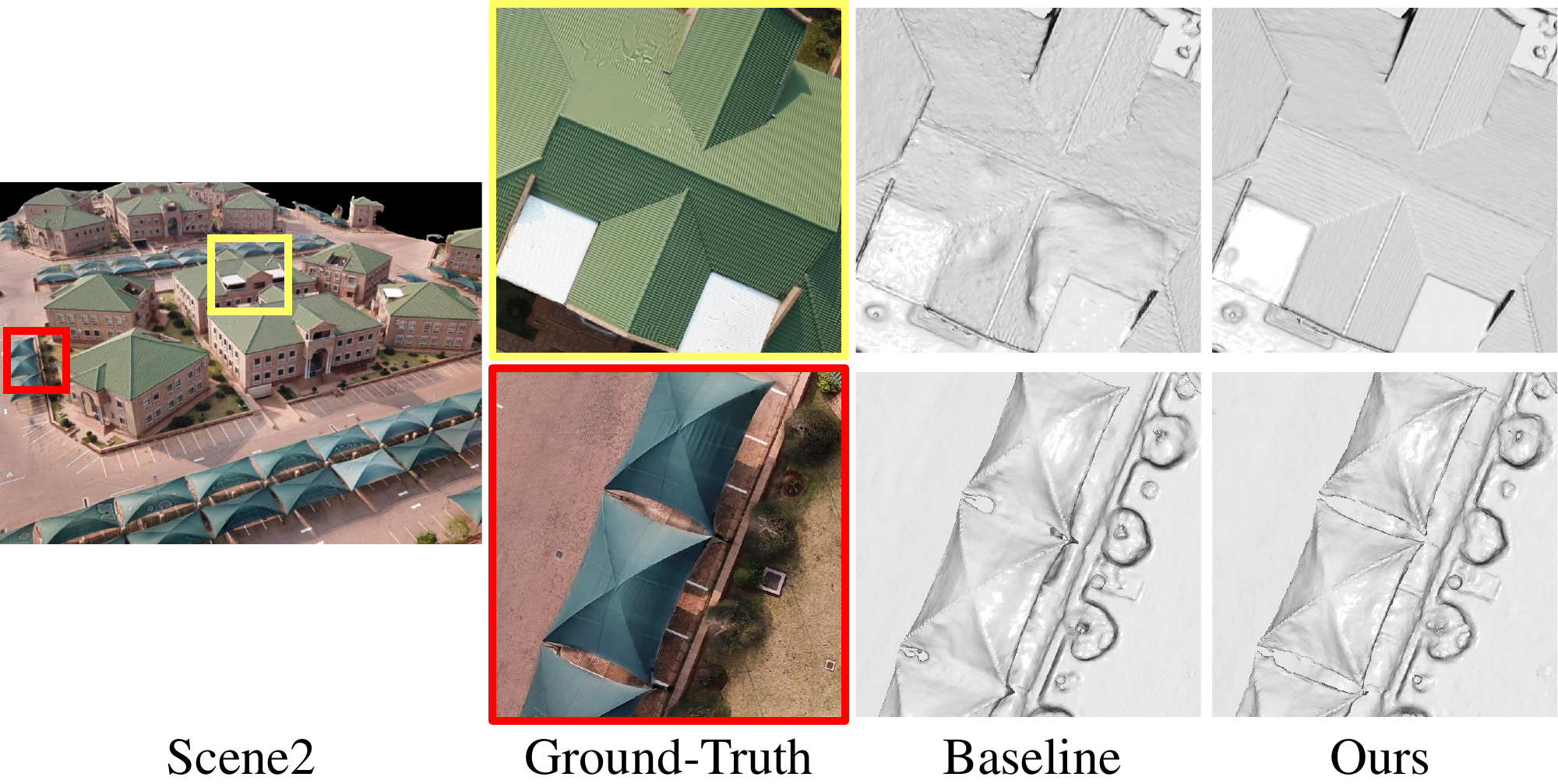}
    \includegraphics[width=0.49\linewidth]{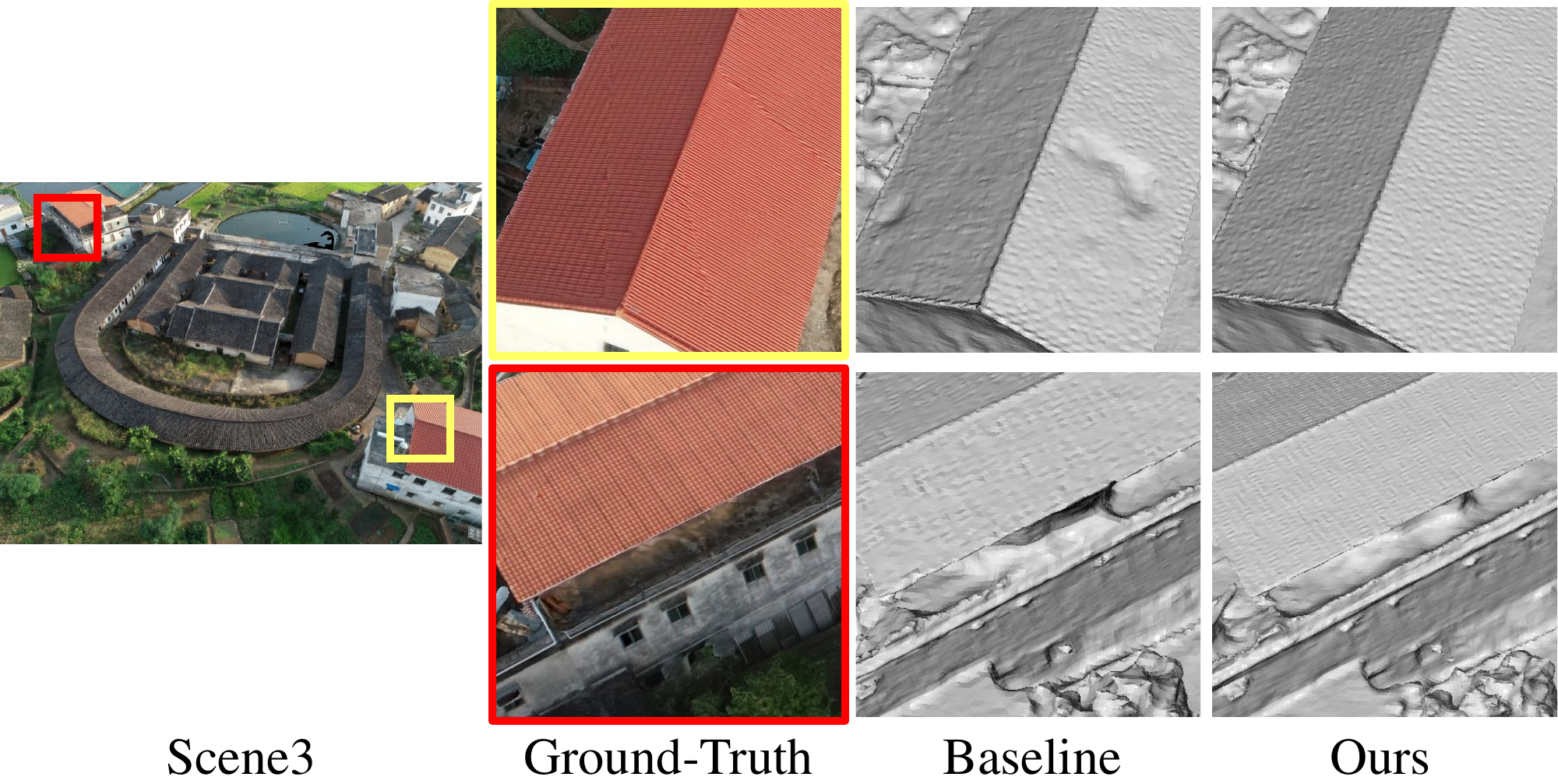}
    \includegraphics[width=0.49\linewidth]{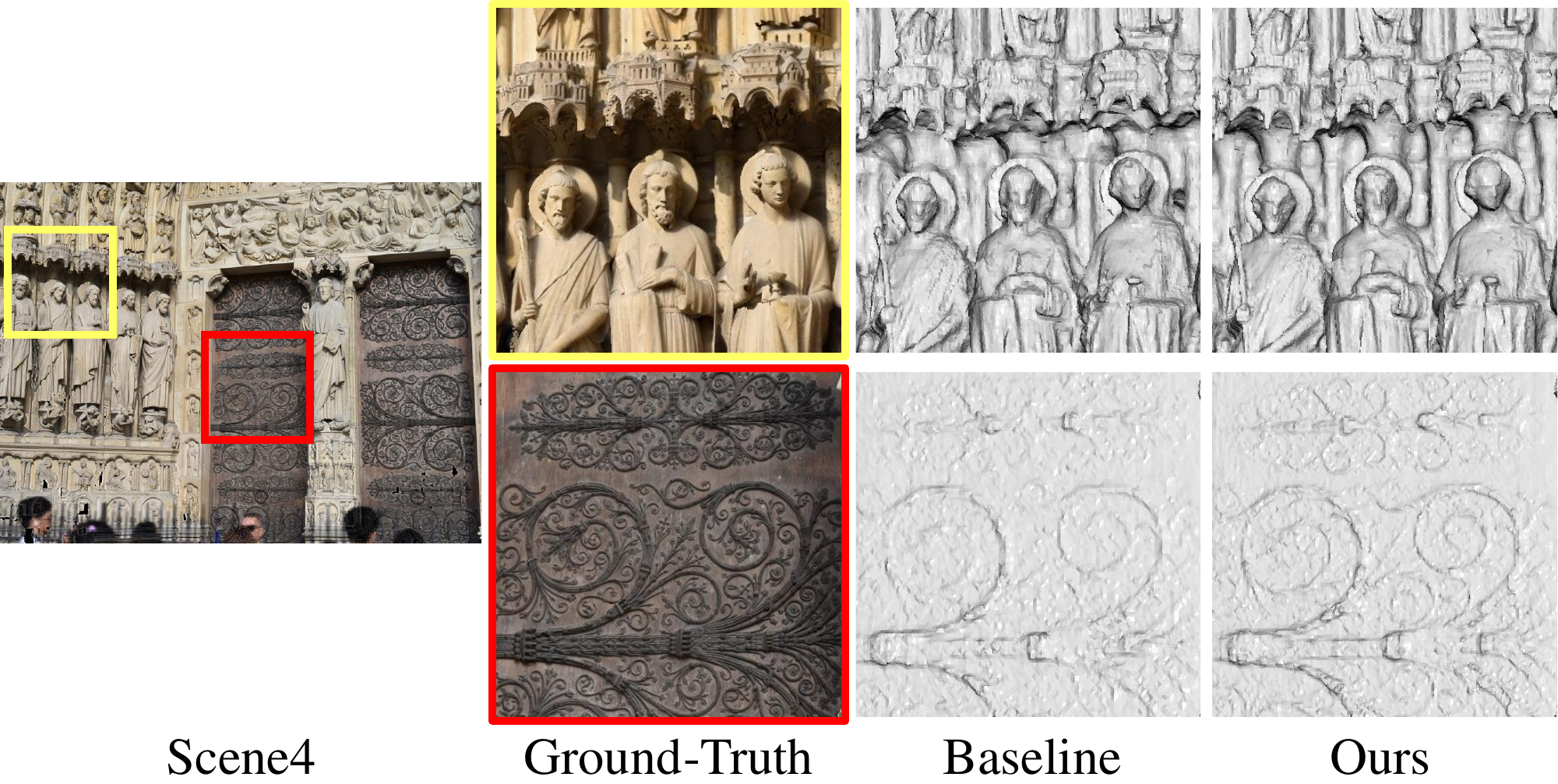}
    \caption{\textbf{Visualization of High-Quality Reconstruction of BlendedMVS Datasets.} By magnifying and comparing local details, our method achieves higher-quality reconstruction and recovers more high-fidelity details.}
  \label{fig:bmvs}
\end{figure*}

\begin{table}
  \centering
  \caption{\textbf{Quantitative Evaluation on the BlendedMVS Datasets.} For F-score, the distance threshold is $5e^{-4}$. The optimal result is indicated in bold.}
  \begin{tabular}{@{}c|lccc@{}}
    \toprule
     & & CD$\downarrow$ ($10^{-6}$) & F-score$\uparrow$  & SDF$\downarrow$ ($10^{-4}$) \\
    \midrule
    \multirow{2}{*}{Scene1} & Baseline & 3.728 & 0.9797 & 7.167 \\
     & 4-nodes & \textbf{2.424} & \textbf{0.9894} & \textbf{4.459}\\
    \midrule
    \multirow{2}{*}{Scene2} & Baseline & 5.923 & 0.9993 & 8.421 \\
     & 4-nodes & \textbf{4.383} & \textbf{0.9999} & \textbf{6.372}\\
    \midrule
    \multirow{2}{*}{Scene3} & Baseline & 7.565 & 0.9974 & 6.107 \\
     & 4-nodes & \textbf{5.847} & \textbf{0.9981} & \textbf{4.893}\\
    \midrule
    \multirow{2}{*}{Scene4} & Baseline & 8.389 & 0.7120 & 9.897 \\
     & 4-nodes & \textbf{6.017} & \textbf{0.7670} & \textbf{6.320}\\
    \bottomrule
  \end{tabular}
  \label{tab:bmvs}
\end{table}

\begin{table}
  \centering
  \caption{\textbf{Quantitative Evaluation on the Human Body.} The optimal result is indicated in bold.}
  \begin{tabular}{@{}lccc@{}}
    \toprule
     & CD$\downarrow$ ($10^{-6}$) & F-score$\uparrow$ (threshold=$5e^{-4}$) & SDF$\downarrow$ ($10^{-4}$) \\
    \midrule
    Baseline & 3.588 & 0.9226 & 8.491 \\
    4-nodes & \textbf{3.403} & \textbf{0.9998} & \textbf{8.450}\\
    \bottomrule
  \end{tabular}
  \label{tab:human}
\end{table}

\begin{figure}
  \centering
  \includegraphics[width=1.0\linewidth]{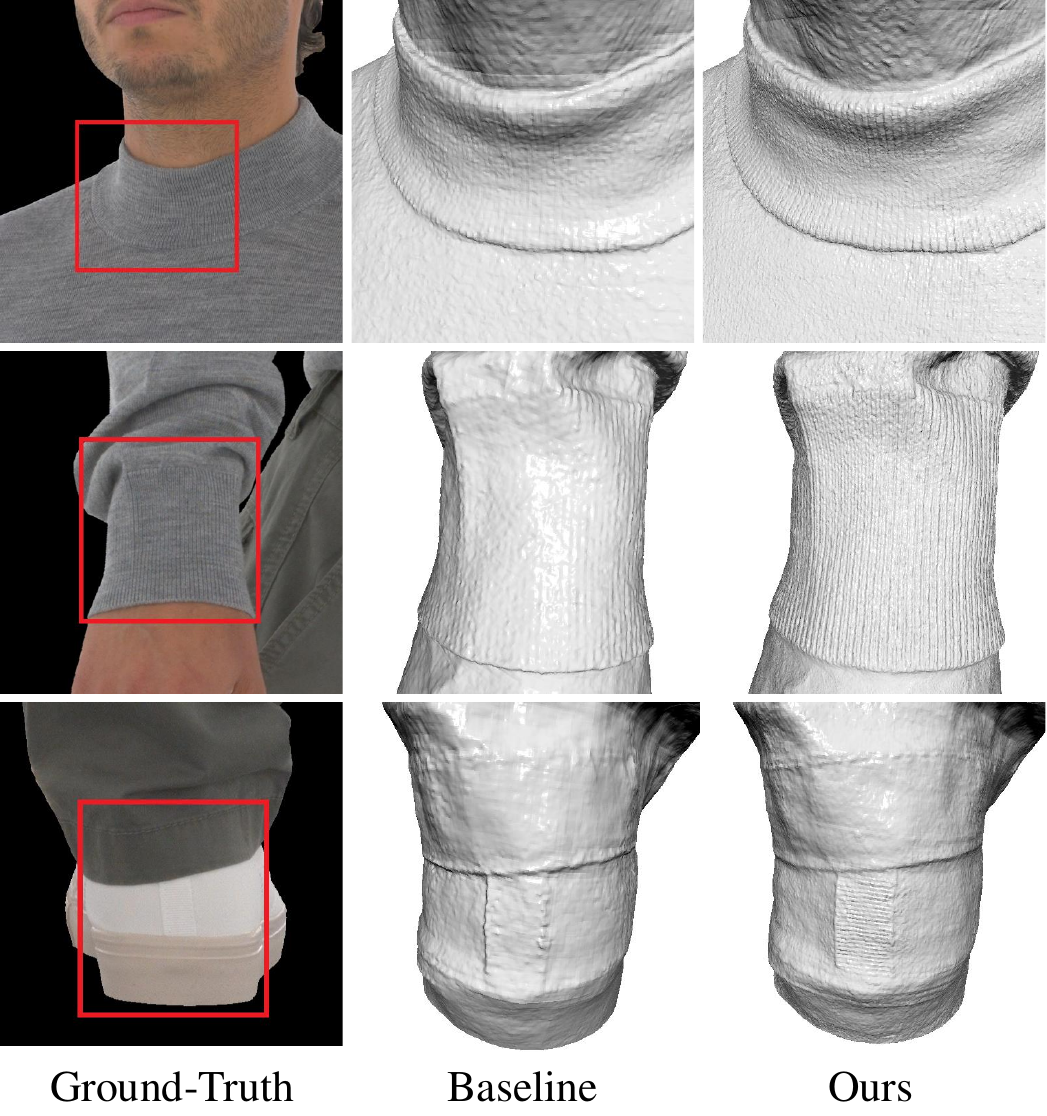}
  \caption{\textbf{Visualization of High-Quality Reconstruction of Human Body.} By the magnified comparison of local areas, our method recovers more details on clothes and shoes than the overall reconstruction.}
  \label{fig:human}
\end{figure}

Table~\ref{tab:lego} and~\ref{tab:jade} show quantitative results from our method and the baseline that uses a single MLP for reconstruction. Figures~\ref{fig:lego} and~\ref{fig:jade} show qualitative visualization of the results. It can be seen that the reconstruction quality improves as the number of nodes increases. Notably, whether it is the concave and convex structures of the knobs in Lego, the outlines of the gears, and the holes in the tires, or the carvings and hollowing out of the jade, they are gradually reconstructed and delineated with increasing detail. Therefore, both quantitative and qualitative results demonstrate that the expressive capability of a single MLP is limited. In comparison, as the number of nodes increases, our method gradually improves the quality of the results and captures the fine details in local regions.

\subsection{High-Quality Reconstruction}
\label{sec:precision}

In the previous section, we verified two fundamental facts and demonstrated the feasibility of our method through designed experiments. In this section, in addition to the high-precision reconstruction of Lego and Jade, we conduct further experiments to demonstrate that our method can achieve high-quality reconstruction. To illustrate the effectiveness of our approach, we compare it with a baseline that reconstructs the whole scene using a single MLP.  Similarly, for a fair comparison, we use the same reconstruction model for the baseline and for our nodes, including, of course, utilizing identical settings for Marching Cubes.

\paraheading{Human body} For the scene of a clothed human body, we divide it into four nodes corresponding to the head, the torso, and two legs, respectively. We use the Bakedangelo model from SDFStudio for reconstruction. This model combines BakedSDF~\cite{bakedsdf} with Neuralangelo~\cite{neuralangelo} to achieve high-quality reconstruction, providing a strong baseline of single-MLP reconstruction for our comparison.

\begin{figure*}
  \centering
    \includegraphics[width=0.56\linewidth]{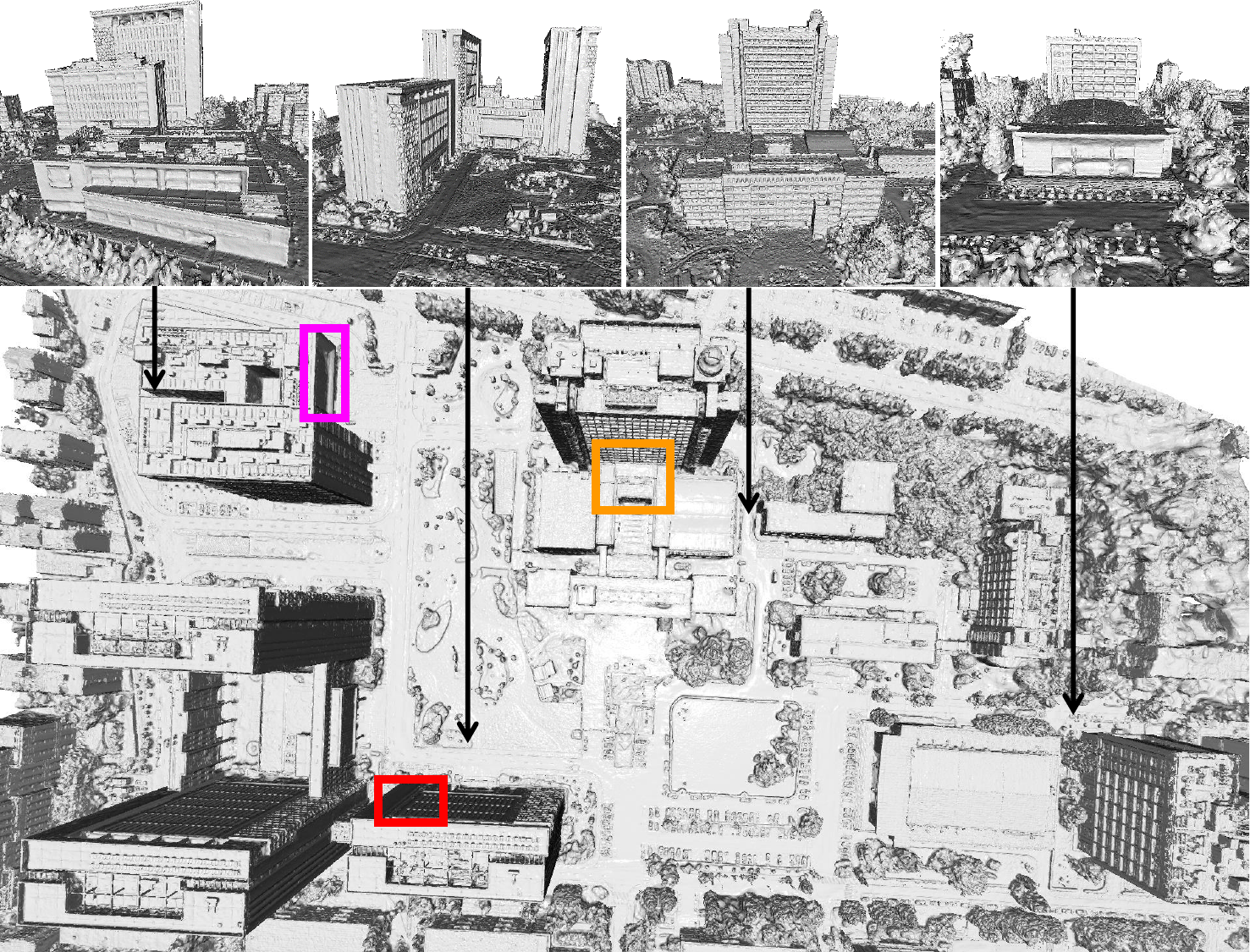}
    \includegraphics[width=0.39\linewidth]{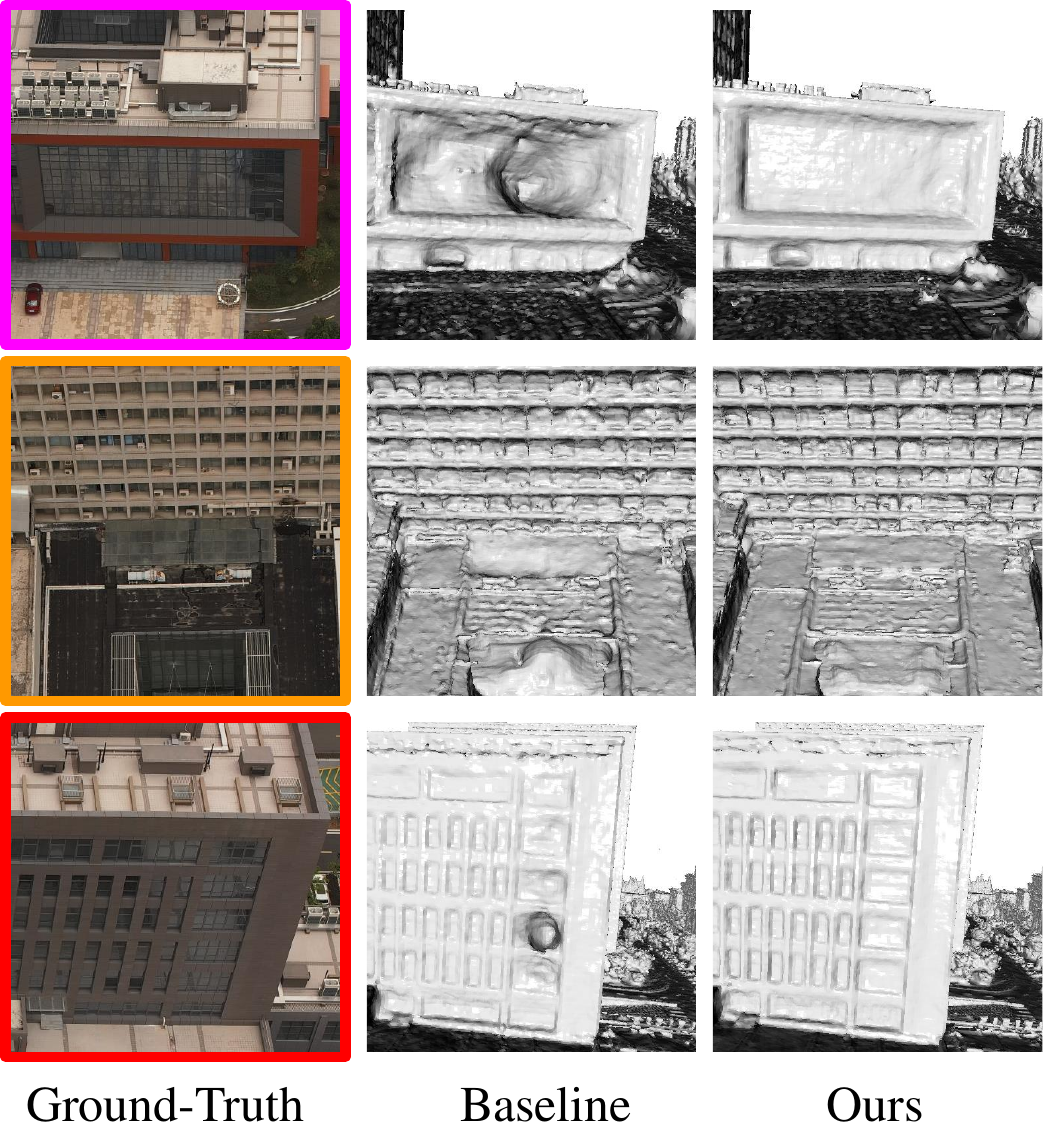}
    \caption{\textbf{Visualization of High-Quality Reconstruction of Sub-Campus with 4 Nodes.} On the left side, we present an overview of the object-based scene Sub-Campus reconstructed using our method, featuring the four objects utilized for node partitioning. A detailed comparison of local results on the right reveals a significant advantage in terms of reconstruction quality for our method.}
  \label{fig:object-based}
\end{figure*}

\begin{table}
  \centering
  \caption{\textbf{The Information of Nodes in Sub-Campus.} We display the number of images for each node, the number of shared images between adjacent node, and the COLMAP running time for each node and baseline (reconstruction as a whole), respectively.}
  \label{tab:object-based}
  \renewcommand
  \arraystretch{1.5}
  \setlength{\tabcolsep}{6pt}
  \begin{tabular}{@{}r|c|c|c|c|c@{}}
    \specialrule{1.5pt}{0pt}{0pt}
     & Node $v_1$ & $v_2$ & $v_3$ & $v_4$ & Baseline \\
    \specialrule{1pt}{0pt}{0pt}
    Total Images & 893 & 561 & 782 & 522 & 2273 \\
    \cline{1-6}
    Shared Images with $v_1$ & - & 140 & 82 & 87 & - \\
    \cline{1-6}
    \multirow{2}{*}{COLMAP Runtime} & 1h47m & 58m & 1h33m & 44m & \multirow{2}{*}{14h25m}\\
    \cline{2-5}
     & \multicolumn{4}{|c|}{Total : 5h8m} & \\
    \specialrule{1.5pt}{0pt}{0pt}
  \end{tabular}
\end{table}

Figure~\ref{fig:human} qualitatively visualizes the results from the baseline and our method.
Our method captures a higher level of detail, particularly the fine stripe patterns on the clothing and the shoes. 
The numerical results in Table~\ref{tab:human} further verify that our approach achieves high-quality reconstruction.

\begin{figure*}
    \centering
    \includegraphics[width=1.0\linewidth]{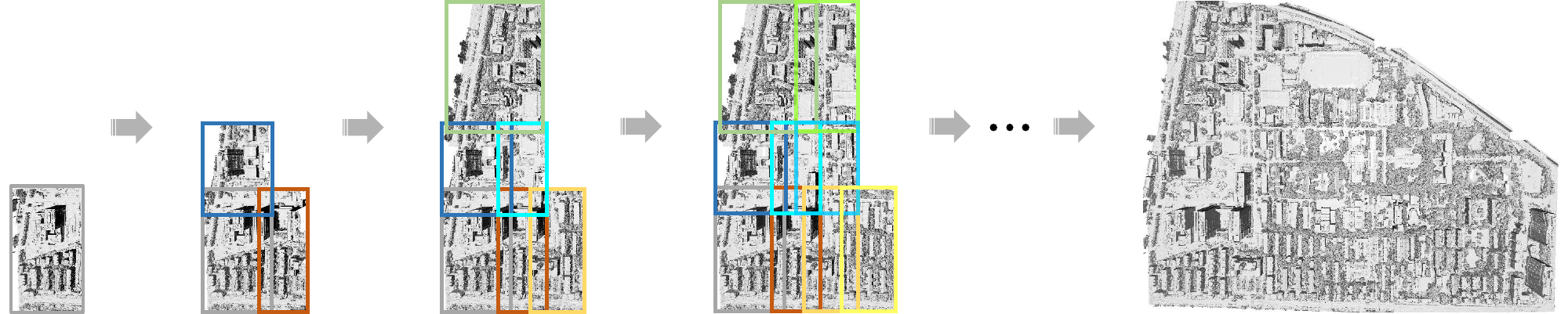}
    \caption{\textbf{Process of data collection and Scalable Scene Reconstruction of Campus Dataset.} The image above illustrates the process of reconstructing the scalable large scene, Campus, 
    based on the direction and extent of data collection.}
  \label{fig:scalable-large-scene-process}
\end{figure*}

\begin{figure*}
    \centering
    \includegraphics[width=1.0\linewidth]{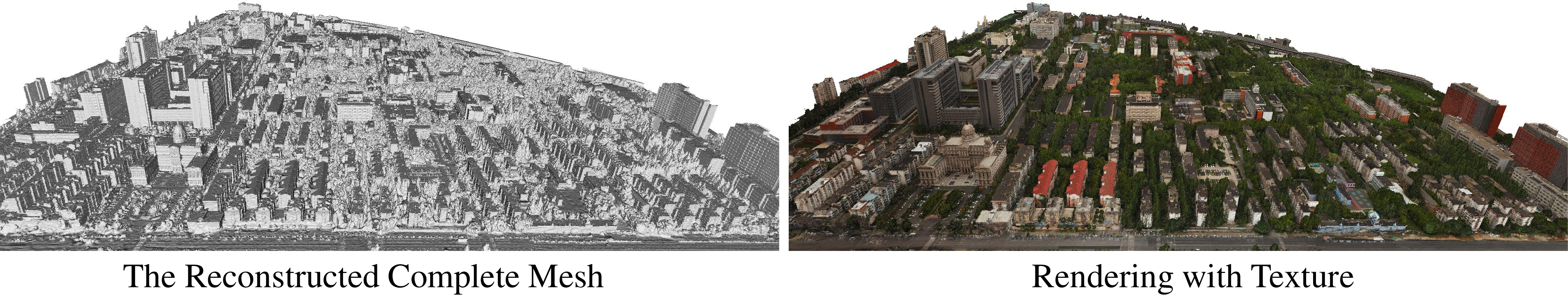}
    \includegraphics[width=1.0\linewidth]{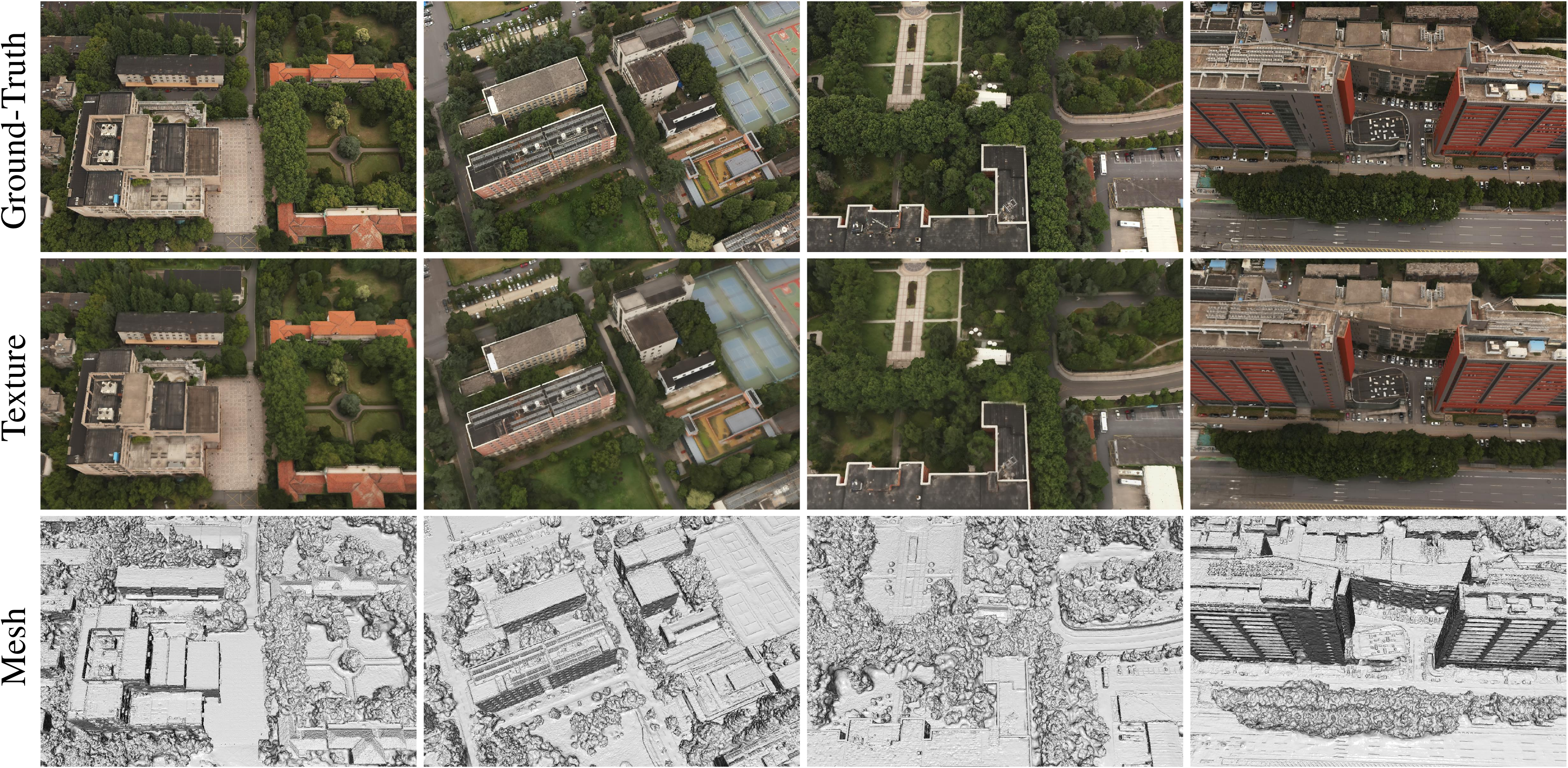}
  \caption{\textbf{Visualization Results of Scalable Scene Reconstruction on Campus Dataset.} A comprehensive view of meshes and textures is presented above. Below is a partial close-up comparison. The top row displays the ground-truth image, the middle row showcases the rendered image using our texture, and the bottom row exhibits the mesh image.}
  \label{fig:scalable-large-scene}
\end{figure*}

\paraheading{BlendedMVS} We select multiple large-scale scenes from the BlendedMVS datasets to further evaluate the advantage of our method in high-quality reconstruction. Due to the relatively narrow range of the z-coordinates in the scene, we conduct partition only along the x-axis and y-axis, dividing the scene into four nodes with overlapping areas. We still utilize the more robust baseline, Bakedangelo, for both the single-MLP reconstruction and each node.

Figure~\ref{fig:bmvs} presents the visual comparison results. It can be seen that, despite the basline is the sota method, our method still achieves higher-quality reconstruction in large-scale scenes, particularly in capturing high-fidelity details such as the flatness of building facades, the integrity of steps and edges, the smoothness of rooftops, and the three-dimensionality of reliefs. Moreover, the numerical comparison results in Table~\ref{tab:bmvs} further ddemonstrate the substantial enhancement in reconstruction accuracy achieved by our method.

\paraheading{Sub-Campus} We partition the scene into four nodes containing different subsets of buildings. We use COLMAP for independent estimation of camera poses for each node, as well as camera pose estimation for the baseline. Table~\ref{tab:object-based} displays the number of images for each node, the number of shared images between adjacent nodes, and the COLMAP running time. The SDF reconstruction is then performed using the Bakedangelo model. 

From Table~\ref{tab:object-based}, it is worth noting that the COLMAP processing takes 14 hours and 25 minutes for the baseline, but only 5 hours and 8 minutes for our approach. This shows our significant advantage in efficiency when dealing with a large amount of images. 
Figure~\ref{fig:object-based} shows some qualitative visualization of the results. Our method correctly captures the building facade shapes, whereas the baseline produces erroneous results in some areas due to limited representational power. Additionally, our result shows clearer windows outlines with improved accuracy. These observations verify the capability of our method in handling large-scale scenes.

\subsection{Scalable Large Scene Reconstruction}
\label{sec:large}

To verify the scalability of our approach, we further test our method on our Campus dataset, which contains 5973 oblique photography images covering an area of about 1200m$\times$800m. At such a large scale, a single-MLP approach becomes impractical in both data processing and SDF-based reconstruction.

According to the flight path employed during the data collection process, we designate the local area collected at the beginning as the starting node. From this node, we gradually expand the collection and nodes to encompass the entire scene. The distribution of nodes and the expansion strategy are depicted in Figure~\ref{fig:scalable-large-scene-process}. Therefore, based on the collection path and data distribution, we partition the whole scene into a total of $25$ nodes, each containing an average of $540$ images. For node pairs with overlapping regions, the approximate count of shared images in their intersection is around $100$. 
Notably, this partition process is automated and does not require manual intervention. At each node, the camera poses are independently estimated using COLMAP, and the SDF is reconstructed using the Bakedangelo model.

\begin{table*}
  \centering
  \caption{\textbf{Rendering Metrics Before and After Registration Optimization.} Assume that node $v_i$ is registered to $v_j$, for shared images between the two nodes, ``\textbf{target}'' represents the rendering metrics achieved after training the neural radiance fields of $v_j$; ``\textbf{initial}'' indicates the rendering metrics in $v_j$ after transforming the camera pose in $v_i$ using the initial registration; ``\textbf{final}'' denotes the rendering metrics obtained after applying the optimized registration. We expect the optimized ``\textbf{final}'' to reach the ``\textbf{target}'' metrics, which means that the registration is aligned. Each set of three columns constitutes a distinct experiment, illustrating the registration of one node to another.}
  \renewcommand\arraystretch{1.2}
  \begin{tabular}{@{}r|ccc|ccc|ccc|ccc|ccc@{}}
    \toprule[1.5pt]
     & \multicolumn{3}{|c|}{Edge 1} 
     & \multicolumn{3}{|c|}{Edge 2} 
     & \multicolumn{3}{|c|}{Edge 3} 
     & \multicolumn{3}{|c|}{Edge 4} 
     & \multicolumn{3}{|c}{Edge 5}\\
    \cline{2-16} 
    \rule{0pt}{12pt}
     & target & initial & final 
     & target & initial & final 
     & target & initial & final
     & target & initial & final
     & target & initial & final \\
    \midrule[1.0pt]
     PSNR$\uparrow$ 
     & 25.94 & 21.22 & 25.35 
     & 27.04 & 22.89 & 26.98 
     & 27.95 & 21.46 & 27.59 
     & 24.52 & 19.28 & 24.21
     & 27.82 & 20.87 & 27.20\\
     SSIM$\uparrow$  
     & 0.803 & 0.518 & 0.781 
     & 0.845 & 0.647 & 0.836 
     & 0.856 & 0.605 & 0.856 
     & 0.839 & 0.485 & 0.824
     & 0.860 & 0.433 & 0.830 \\
     LPIPS$\downarrow$ 
     & 0.096 & 0.116 & 0.096 
     & 0.078 & 0.095 & 0.078 
     & 0.069 & 0.091 & 0.069  
     & 0.075 & 0.109 & 0.075
     & 0.064 & 0.101 & 0.064 \\
    \bottomrule[1.5pt]
  \end{tabular}
  \label{tab:regix}
\end{table*}

\begin{figure*}
  \centering
  \includegraphics[width=1.0\linewidth]{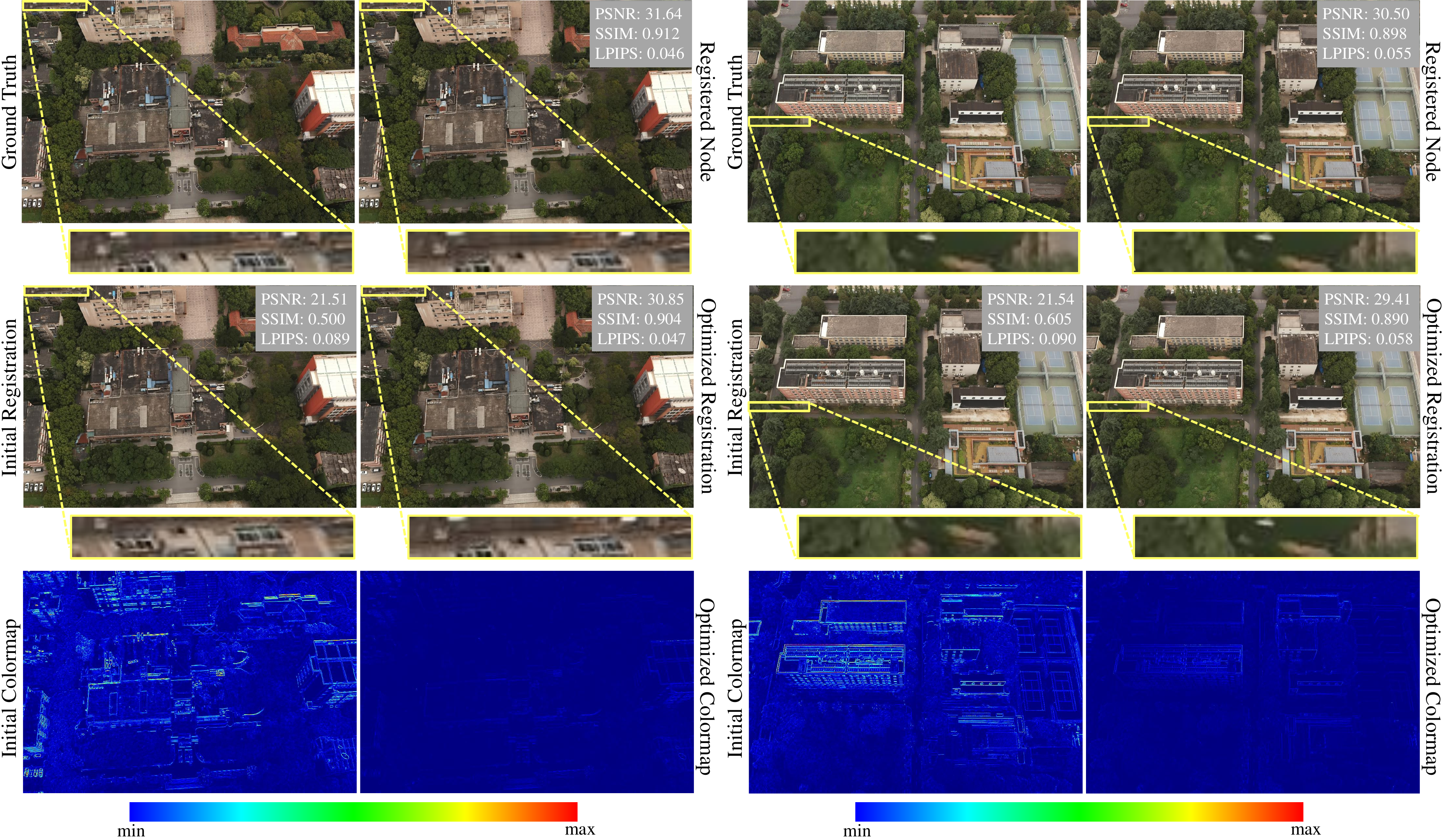}
  \caption{\textbf{Rendering Evaluation for Optimization of Registration.} We present rendering visualizations and metrics for the registered node, initial registration, optimized registration, and a colormap indicating errors relative to supervision (registered node). The partial magnification and rendering metrics reveal significant deviations in the initial registration, as clearly depicted in the figure as an upward bias relative to the reference. Through our optimization method, we effectively rectify these errors.}
  \label{fig:rigs-color}
\end{figure*}

We use the coordinate system of the starting node as the global coordinate system and propagate it along the minimum spanning tree of nodes. Consequently, the registration process involves $24$ edges. Due to space limitations, Table~\ref{tab:regix} only presents the rendering metrics of five edges before and after registration optimization. 
It can be seen that the rendering metrics of the initial registration (the ``initial'' column) are significantly poorer compared to the rendering metrics of the registered node (the ``target'' column), indicating a considerably larger error in the initial registration. By applying our method, we note a substantial improvement in the rendering metrics after optimization (the "final" column), which essentially approaches the metrics of the registered node. To more clearly demonstrate the necessity of registration optimization, we illustrate the changes in the mesh of two adjacent nodes before and after registration optimization in Figure~\ref{fig:ris-ori-opt}. The initial registration pose exhibits significant abnormal deviation and misalignment in the relative positions of the meshes, and after optimization, the two meshes achieve perfect alignment. Furthermore, we present the rendering-based visual comparison before and after registration optimization in Figure~\ref{fig:rigs-color}. The error color map shows a significant misalignment after the initial registration, which is substantially reduced after the registration optimization. Overall, the comparison of all these numerical and visual results demonstrates the necessity and effectiveness of our registration optimization method.

After registration optimization, we conduct blending for all the SDFs of the 25 nodes according to the method described in Section~\ref{sec:link}. Subsequently, we employed Marching Cubes to extract the mesh of the entire scene. The complete mesh of the reconstructed scene and enlarged details in localized areas are depicted in Figure~\ref{fig:scalable-large-scene}. It can be seen that our method faithfully reconstructs the geometry as well as the texture of the whole scene. This demonstrates the effectiveness of our method in handling urban-scale scenes that are intractable for existing single-MLP reconstruction methods.

\subsection{Additional Application Results}
\label{sec:tex-and-edit}

\begin{figure}
  \centering
  \includegraphics[width=\linewidth]{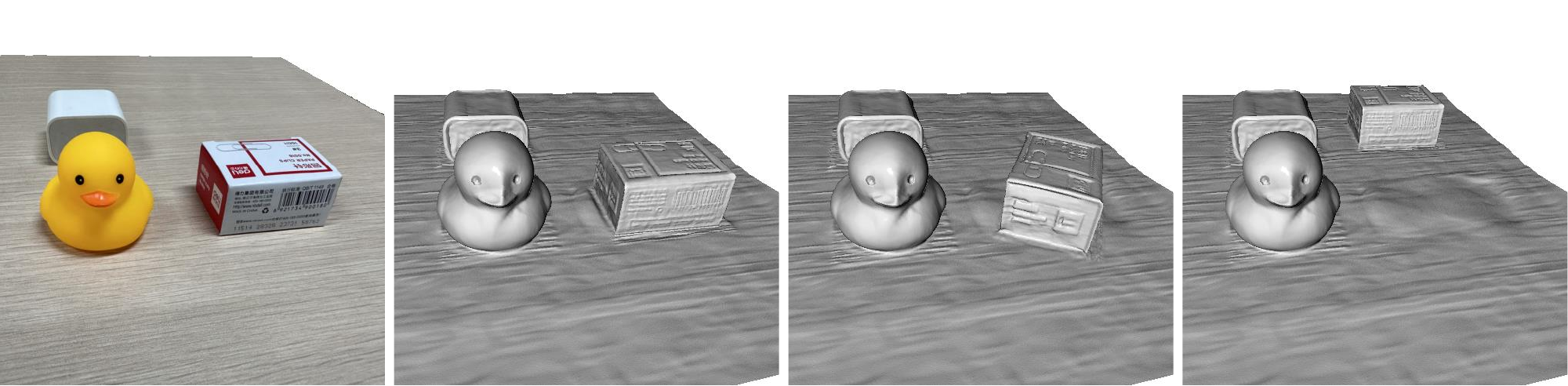}
  \caption{\textbf{Editing Results.} From left to right: RGB images, original mesh, edited mesh examples 1 and 2.}
  \label{fig:editing}
\end{figure}

In addition to the experiments and results shown above, we will show two applications proposed in Section~\ref{sec:application}. Firstly, we have generated texture maps for the previously reconstructed scalable large scene, as illustrated in Figure~\ref{fig:scalable-large-scene}. We utilized Xatlas\footnote{\url{https://github.com/mworchel/xatlas-python}} for parameterization and then employed the surface rendering ``SoftPhongShader'' model with ``AmbientLights'' provided by PyTorch3D~\cite{pytorch3d} for rendering. More details regarding rasterization settings and shader parameter configuration can be found in the appendix. The results demonstrate a high level of fidelity, which further substantiates the practical value of our method. Secondly, using the NeuS model, we apply rotation and translation to the registration transformation of a node within the scene to achieve the effect of editing, as illustrated in Figure~\ref{fig:editing}. This once again illustrates the versatility and practicality of our method.

\subsection{Ablation Study}
\label{seca:ablation-study}

We conduct an ablation study from three aspects: (1) examining the impact of registration optimization, (2) comparing the results with and without SDF blending, and (3) exploring the influence of network parameter count on model performance.

\paraheading{Registration optimization} We first focus on the impact of registration optimization on the accuracy of the reconstruction. We present a visual comparison before and after registration optimization at the mesh level in Figure~\ref{fig:ris-ori-opt}, a rendering-level visual comparison in Figure~\ref{fig:rigs-color}, and more numerical comparison of additional rendering metrics in Table~\ref{tab:regix}. These results demonstrate significant errors in the initial registration, highlighting the necessity and effectiveness of the registration optimization method we proposed.

\begin{figure}
  \centering
  \includegraphics[width=1.0\linewidth]{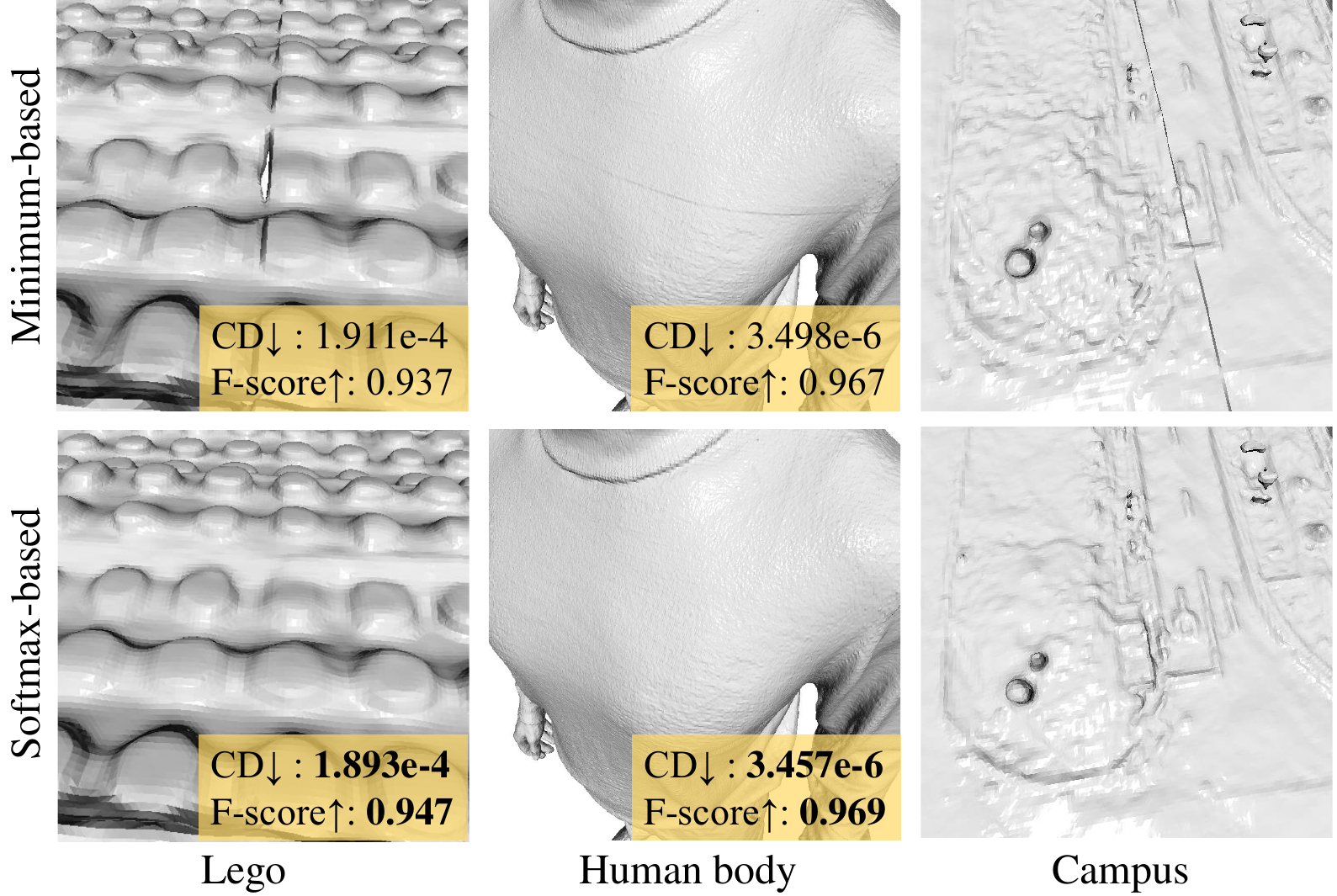}
  \caption{\textbf{Ablation Study of SDF Blending.} Quantitatively evaluate the precision of the mesh after minimum-based and softmax-based SDF blending in the common area respectively. Clearly, our proposed softmax-based blending not only eliminates the seam problem but also maintains the accuracy of the reconstruction.}
  \label{fig:ablation-blending}
\end{figure}

\paraheading{SDF blending} The comparison in Figure~\ref{fig:blending} shows that the minimum-based blending, according to the SDF definition, results in visible seams, whereas our proposed softmax-based blending method effectively eliminates this issue. The presence of seams highlights the necessity for SDF blending. At the same time, we quantitatively evaluate the reconstruction quality after blending in the common area, as shown in Figure~\ref{fig:ablation-blending}. It is evident that softmax-based blending not only effectively resolves the seam problem, but also addresses the issue of poor reconstruction quality at the boundaries of each node, (the boundary of a node being a relatively internal position compared to adjacent nodes). Consequently, the reconstruction quality is not compromised due to SDF blending.

\begin{table}
  \centering
  \caption{\textbf{Ablation Study on Parameter Count.} For the F-score, the distance threshold is $5e^{-4}$. ``$\nicefrac{1}{4}$ layers'' and ``$\nicefrac{1}{4}$ hidden'' represents reducing the baseline network to one-fourth of the original size and then applying our method. ``$\times 4$ iterations'' means increasing the number of iterations. ``$\times 4$ layers'' and ``$\times 4$ hidden'' indicate quadrupling the baseline network and correspondingly the number of iterations. The optimal results are indicated in bold.}
  \begin{tabular}{@{}lccc@{}}
    \toprule
     & CD$\downarrow$ ($10^{-4}$) & F-score$\uparrow$ & SDF$\downarrow$ ($10^{-3}$) \\
    \midrule
    $\nicefrac{1}{4}$ layers & 4.953 & 0.8524 & 8.632 \\
    $\nicefrac{1}{4}$ hidden & 5.777 & 0.8115 & 8.936 \\
    \midrule
    Baseline & 5.209 & 0.8445 & 8.779 \\
    \midrule
    $\times 4$ iterations & 5.048 & 0.8547 & 8.252 \\
    $\times 4$ layers & 5.135 & 0.8595 & 8.452 \\
    $\times 4$ hidden & 5.207 & 0.8551 & 8.564 \\
    \midrule
    Ours 4-nodes & \textbf{3.877} & \textbf{0.8826} & \textbf{7.054} \\
    \bottomrule
  \end{tabular}
  \label{tab:mlp-lego}
\end{table}

\begin{figure*}
  \centering
  \includegraphics[width=1.0\linewidth]{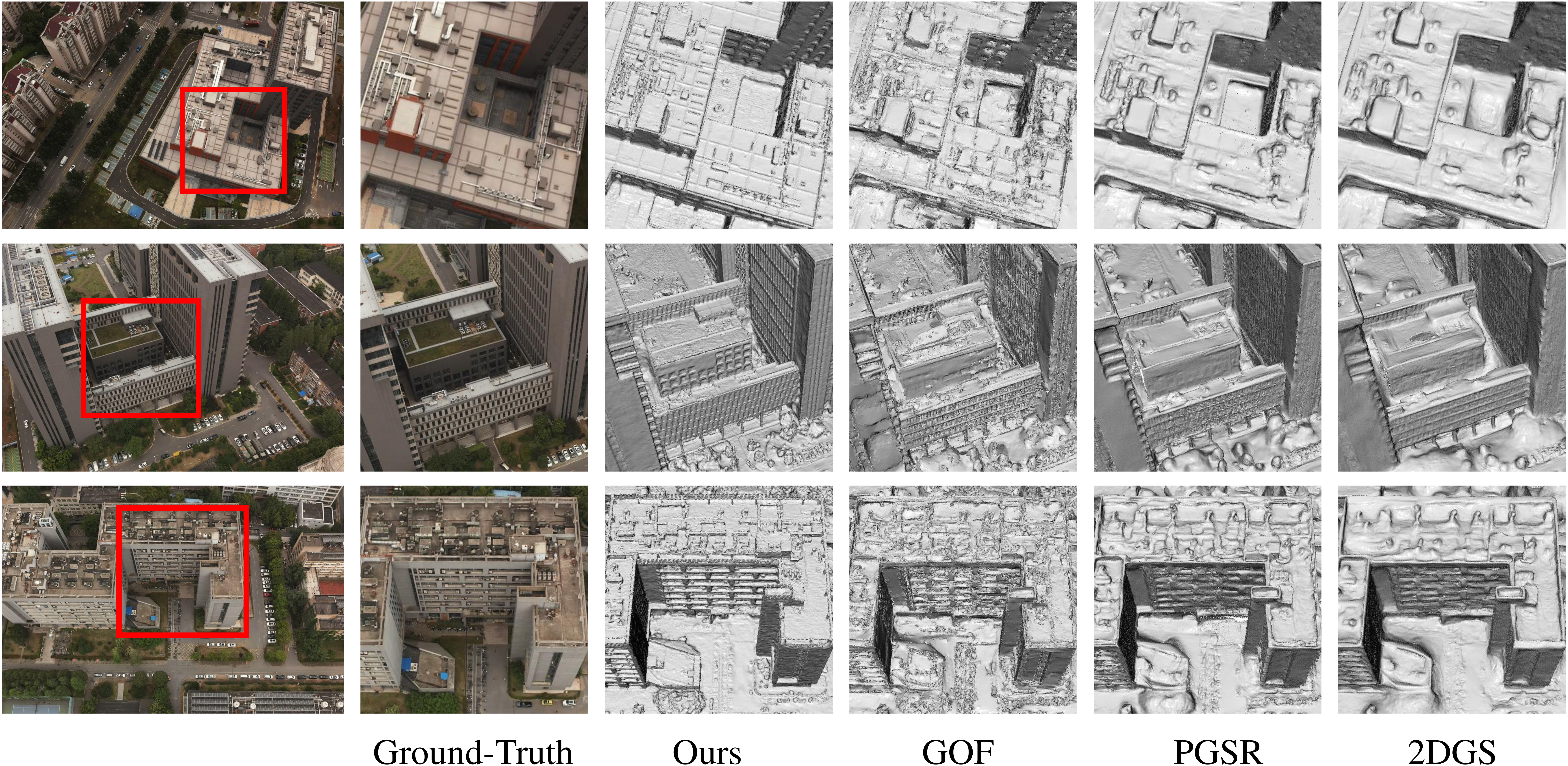}
  \caption{\textbf{Visualization Results of Sub-Campus Dataset Compared with Other Representation.} We compare our method with some novel and state-of-the-art geometric surface reconstruction methods based on 3DGS~\cite{3dgs}, including Gaussian Opacity Fields (GOF)~\cite{gof}, PGSR~\cite{pgsr}, and 2D Gaussian Splatting (2DGS)~\cite{2dgs}. It can be observed that our method has better reconstruction ability, and even the current novel 3DGS-based surface reconstruction methods also face limitations in expressive capability when dealing with large-scale scene.}
  \label{fig:compare_gauss}
\end{figure*}

\paraheading{Parameter count}
The expressiveness of the implicit representation is directly related to the parameter count, which depends on the network's layer depth and the number of neurons per layer. To ensure a fair comparison and further demonstrate the effectiveness of our method, we conduct an ablation study on the number of network parameters. We utilize the high-precision Lego reconstruction based on 4 nodes in Section~\ref{sec:validation-of-framework-effectiveness} for the comparison. This study is divided into two aspects. First, to verify that simply increasing the number of parameters and training iterations for a single network cannot significantly improve the precision, we quadrupled the parameter count of the baseline single network and correspondingly quadrupled the number of iterations. Second, to show that our method can improve the reconstruction accuracy while keeping the number of parameters fixed, we reduced the baseline model to one-fourth of its original size and then combined t with our method. Notably, both increasing and decreasing parameters involve adjusting the number of network layers and the number of neurons in each hidden layer.

Table~\ref{tab:mlp-lego} presents the performance metrics of different models. The term ``Baseline'' refers to reconstruction using the original single network model. ``$\times4$ iterations'' indicates that the number of training iterations is quadrupled compared to the baseline, while ``$\times4$ layers'' and ``$\times4$ hidden'' respectively refer to the cases where the number of network layers and the number of neurons per layer in the baseline are each quadrupled, and the corresponding number of iterations are also quadrupled. It can be observed that simply increasing the number of parameters and training iterations for a single network yields only a modest accuracy improvement; however, the gains are minimal relative to the substantial computational and time costs incurred. This highlights that, with equivalent parameter counts, our proposed method is both effective and efficient for high-precision reconstruction. It is important to note that reducing the number of neurons in each layer of the baseline network to ${1}/{4}$ of the original number and using four such networks for reconstruction, as shown in the row ``$\nicefrac{1}{4}$ hidden'', leads to poorer reconstruction performance compared with the original ``Baseline'' row. This is because reducing the size of the hidden layer diminishes the modelling ability of the network output features. On the other hand, reducing the number of layers to $1/4$ of the original baseline network and then reconstructing with four such networks through our method, as shown in the row ``$\nicefrac{1}{4}$ layers'', results in better performance than the original network, indirectly validating the effectiveness of our divide-and-conquer strategy.

\subsection{Comparison with Other Representations}
\label{seca:com_gauss}

Recently, 3D Gauss Splatting (3DGS)~\cite{3dgs} has become a highly promising representation for 3D scenes due to its excellent capability for color field expression. To further demonstrate the high-precision reconstruction capabilities of our method, we compare our approach with some state-of-the-art geometric surface reconstruction methods based on 3DGS, including Gaussian Opacity Fields (GOF)~\cite{gof}, PGSR~\cite{pgsr}, and 2D Gaussian Splatting (2DGS)~\cite{2dgs}. Specifically, we perform comparative tests on the Sub-Campus dataset, using parameter settings from the original papers of these methods. Notably, these 3DGS-based methods still reconstruct the entire scene using a single model. Therefore, to ensure a fair comparison, we increase the training iterations for each model to 200K, allowing full convergence and optimal training.


Figure~\ref{fig:compare_gauss} presents the qualitative visualization comparison results. We can see that for large-scale scenes with sufficiently high complexity, these 3DGS-based methods still face limitations in expressive capability. For example, they cannot accurately reconstruct the flat surfaces of buildings or the refined contours of windows. In contrast, our method achieves more accurate and detailed surface reconstruction. This further emphasizes the effectiveness effectiveness and high accuracy of our divide-and-conquer strategy.

\section{Conclusion and Discussion}
\label{sec:conclusion}

We introduced Neural SDF-Graph, a neural implicit surface reconstruction framework for scalable and high-quality reconstruction, editing, and more. Leveraging the locality of SDF, we represent individual components of a scene as graph nodes, where the SDF for each node is first reconstructed separately and then aligned via registration to derive a global SDF for the whole scene. This enables independent and flexible operations at different nodes, significantly enhancing the scalability and accuracy of SDF-based reconstruction.

Our approach can be improved from multiple aspects. First, our registration optimization is currently performed with an off-the-shelf solver. It is possible to design a specialized solver to boost its efficiency. Additionally, existing neural implicit reconstruction methods assume watertight surfaces, which may not be true for each node's shape. This can lead to irregular iso-surface shapes around the node boundary. Although our SDF blending can mitigate this issue and produce a smooth global SDF, it would be interesting to develop a reconstruction method for non-watertight surfaces to eliminate this problem.
Finally, although we only apply our divide-and-conquer approach to the reconstruction of 3D scenes represented with SDFs in this paper, our idea is also applicable to other 3D representations. In the future, we plan to investigate its extension to explicit 3D representations such as 3DGS.

\ifCLASSOPTIONcompsoc
  \section*{Acknowledgments}
\else
  \section*{Acknowledgment}
\fi

This research was supported by the National Natural Science Foundation of China (No.62441224, No.62272433), the Fundamental Research Funds for the Central Universities (No.WK0010000090), and USTC Fellowship (No.S19582024). The numerical calculations in this paper have been done on the supercomputing system in the Supercomputing Center of University of Science and Technology of China.

\bibliographystyle{IEEEtran}

\bibliography{ref}

\begin{IEEEbiography}
[{\includegraphics[width=1in]{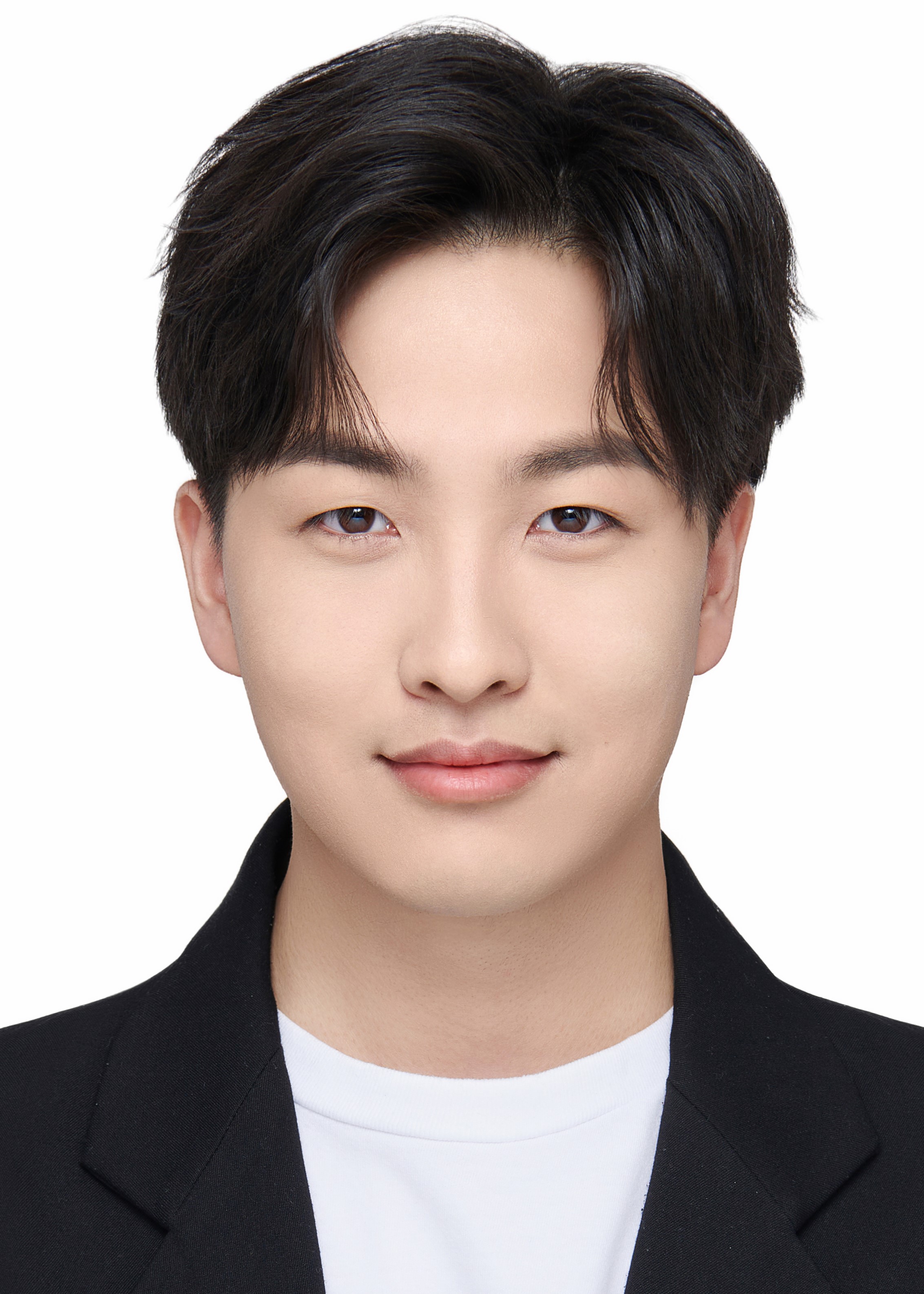}}]
{Leyuan Yang} is currently working as a master student in the School of Mathematical Science, University of Science and Technology of China. His research interests include computer vision, 3D vision and large-scale scene reconstruction.
\end{IEEEbiography}
\begin{IEEEbiography}
[{\includegraphics[width=1in]{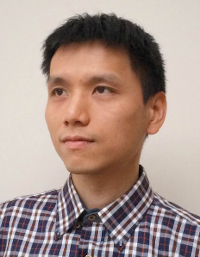}}]
{Bailin Deng} is a Senior Lecturer in the School of Computer Science and Informatics at Cardiff University. He received the BEng degree in computer software (2005) and the MSc degree in computer science (2008) from Tsinghua University (China), and the PhD degree in technical mathematics (2011) from Vienna University of Technology (Austria). His research interests include geometry processing, numerical optimization, computational design, and digital fabrication. He is a member of the IEEE, and an associate editor of IEEE Computer Graphics and Applications.
\end{IEEEbiography}
\begin{IEEEbiography}
[{\includegraphics[width=1in]{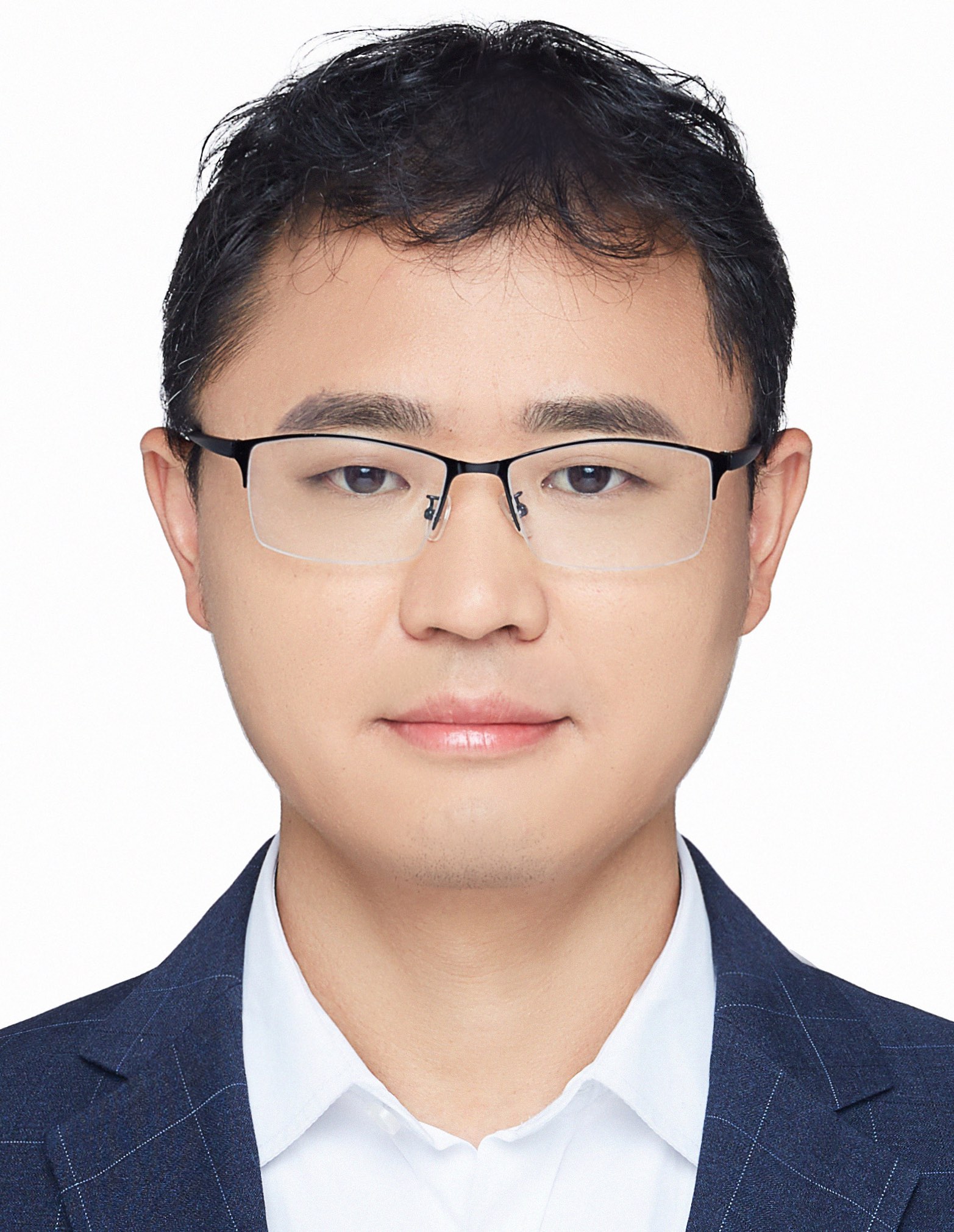}}]
{Juyong Zhang} is a professor in the School of Mathematical Sciences at University of Science and Technology of China. He received the BS degree from the University of Science and Technology of China in 2006, and the PhD degree from Nanyang Technological University, Singapore. His research interests include computer graphics, computer vision, and numerical optimization. He is an associate editor of IEEE Transactions on Multimedia.
\end{IEEEbiography}
\clearpage

\appendices
\section{Experimental Configuration and Details}

We will provide details on the configurations of our experiments, including network structure, parameter settings.

\subsection{SDF Method and Configuration}

We first present the network structure and some basic parameter settings for the two selected SDF-based methods, NeuS and Bakedangelo. For additional details, please refer to the original project homepage and paper.

\subsubsection{NeuS}
\paraheading{Loss}
NeuS utilizes only photometric loss~\cite{nerf}, Eikonal loss~\cite{eikonal} and mask loss. We set the loss weights as $\lambda_{color}=1.0$, $\lambda_{eikonal}=0.1$, and $\lambda_{mask}=0.1$.

\paraheading{Network architecture}
We retain the original network architecture of NeuS, consisting of two MLPs to encode SDF and color respectively. The signed distance function $f$ is modeled by an MLP that consists of $8$ hidden layers with hidden size of $256$. The activation function for all hidden layers is Softplus with $\beta=100$, and a skip connection is used to connect the input with the output of the fourth layer. The function $\boldsymbol{c}$ for color prediction is modeled by an MLP with $4$ hidden layers with size of $256$, which takes not only the spatial location $\boldsymbol{p}$ as inputs but also the view direction $\boldsymbol{d}$, the normal vector of SDF $\nabla f(\boldsymbol{p})$, and a $256$-dimensional feature vector from the SDF MLP. For additional information on the network structure and related details of NeuS, please consult the original paper.

\paraheading{Training details}
The neural networks of NeuS are trained using the Adam optimizer. The learning rate is first linearly warmed up from $0$ to $5\times10^{-4}$ in the first 5K iterations, and then controlled by the cosine decay schedule to the minimum learning rate of $2.5\times10^{-5}$. We use $2048$ rays in each iteration. The number of iterations is determined based on various experimental objectives and datasets, so we will elaborate on the specific experimental details below.

\subsubsection{Bakedangelo}
\paraheading{Loss}
In addition to the fundamental photometric loss, Eikonal loss and mask loss, Bakedangelo incorporates curvature loss~\cite{neuralangelo} and interlevel loss~\cite{mipnerf360, zipnerf}. We set the loss weights as $\lambda_{color}=1.0$, $\lambda_{eikonal}=0.01$, $\lambda_{mask}=0.1$, $\lambda_{curvature}=0.0005$, and $\lambda_{interlevel}=1.0$.

\paraheading{Network architecture}
We also retain the original network structure of Bakedangelo. The signed distance function $f$ is modeled by an MLP that consists of only $1$ hidden layers with a size of $256$, and the function of color prediction $\boldsymbol{c}$ is modeled by an MLP with $4$ hidden layers with a size of $256$. Simultaneously, the model leverages a proposal MLP~\cite{mipnerf360} for guiding the sampling process. Building upon Neuralangelo~\cite{neuralangelo}, the Bakedangelo model progressively activates the Hash encoding level.

\paraheading{Training details}
The neural networks are trained using the Adam optimizer. The bias for initializing the SDF field is $bias=0.1$. The learning rate for the SDF network is first linearly warmed up from $0$ to $0.001$ in the first 5K iterations, and then controlled by the exponential decay schedule with milestones set at 600K and 800K iterations. The learning rate for the Proposal MLP is $0.01$. The number of Hash encoding layers is $level\_init=8$. In each iteration we use $1024$ rays. We basically maintained the initial settings of this model in SDFStudio. Similarly, the number of iterations relies on various experimental goals and datasets, so a detailed explanation will be provided below. For other detailed parameter settings, please refer to the original project.

\subsection{Implementation Details}

In this part, we elaborate on the experimental details for our reconstruction. 

\paraheading{Lego} 
We utilize the NeuS model for reconstruction. Considering the total number of rays is $100\times800\times800=6.4\times10^{7}$, the number of rays in each iteration is $2048$, and we assume an average training time is $10$ per ray, that is, $10\approx6.4\times10^{7}\times \text{\#{iterations}}/2048$, so the total number of iterations is set to 310K.

\paraheading{Jade} 
We still use the NeuS model for reconstruction. We directly apply original experimental configuration of NeuS for the BlendedMVS datasets, that is, 512 rays per iteration, and the total number of iterations is 300K.

\paraheading{Human body}
We employ the Bakedangelo model. The iteration count for the baseline is set to 3000K, while the iteration count for each node is set to 750K.

\paraheading{BlendedMVS}
We employ the Bakedangelo model for reconstruction. The iteration count for the baseline is set to 400K, while the iteration count for each node is set to 100K.

\paraheading{Sub-Campus}
We employ the Bakedangelo method for reconstruction in both the baseline (reconstructing as a whole) and each node. The iteration count for the baseline is set to 4000K, while the iteration count for each node is set to 1000K. 

\paraheading{Campus}
We still employ the Bakedangelo method, and the number of training iterations for each node is 1000K.

\subsection{Surface Rendering Configuration}
For the experiments in Section~\ref{sec:application} to generate textures based on our method, we used the "SoftPhongShader" surface rendering model provided by PyTorch3D. The rasterization configuration parameters are: the distance used to expand the face bounding boxes for rasterization $blur\_radius=0.0$, number of faces to keep track of per pixel $faces\_per\_pixel=1$, size of bins to use for coarse-to-fine rasterization $bin\_size=125$, whether to apply perspective correction when
computing barycentric coordinates for pixels $perspective\_correct=True$, whether to correct a location outside the face to a position on the edge of the face $clip\_barycentric\_coords=False$, whether to only rasterize mesh faces which are visible to the camera $cull\_backfaces=False$. As for the lighting, we select the ambient light model ``AmbientLights'', and the parameter is $ambient\_color=(0.95, 0.95, 0.95)$. Additionally, for "SoftPhongShader" , the parameter for controling the width of the sigmoid function used to calculate the 2D distance based probability is $blend\_params=BlendParams(sigma=10^{-5})$.

\end{document}